\documentclass[10pt,twocolumn,letterpaper]{article}

\usepackage{iccv}
\usepackage{times}
\usepackage{epsfig}
\usepackage{graphicx}
\usepackage{amsmath}
\usepackage{amssymb}

\usepackage{booktabs}

\usepackage[utf8]{inputenc} 
\usepackage[T1]{fontenc}    
\usepackage{url}            
\usepackage{booktabs}       
\usepackage{amsfonts}       
\usepackage{nicefrac}       
\usepackage{microtype}      
\usepackage[table]{xcolor}         

\usepackage{amsmath}
\usepackage[numbers]{natbib}
\usepackage{xspace}
\usepackage{graphicx}
\usepackage[export]{adjustbox}
\usepackage{enumitem}
\usepackage{algorithm}
\usepackage[noend]{algpseudocode}
\usepackage{verbatim}
\usepackage{subcaption}
\usepackage{multirow}
\usepackage{hhline}
\usepackage{pifont}  
\usepackage{wrapfig}

\usepackage{nicematrix}
\usepackage{tabularx}
\usepackage{colortbl, hhline}
\usepackage{todonotes}

\usepackage[pagebackref=true,breaklinks=true,letterpaper=true,colorlinks,bookmarks=false]{hyperref}

\usepackage{cleveref}
\crefname{section}{Sec.}{Secs.}
\Crefname{section}{Section}{Sections}
\Crefname{table}{Table}{Tables}
\crefname{table}{Tab.}{Tabs.}

\crefformat{section}{\S#2#1#3}
\crefformat{subsection}{\S#2#1#3}
\crefformat{subsubsection}{\S#2#1#3}
\crefrangeformat{section}{\S\S#3#1#4 to~#5#2#6}
\crefmultiformat{section}{\S\S#2#1#3}{ and~#2#1#3}{, #2#1#3}{ and~#2#1#3}
\Crefformat{figure}{#2Fig.~#1#3}
\Crefmultiformat{figure}{Figs.~#2#1#3}{ and~#2#1#3}{, #2#1#3}{ and~#2#1#3}
\Crefformat{table}{#2Tab.~#1#3}
\Crefmultiformat{table}{Tabs.~#2#1#3}{ and~#2#1#3}{, #2#1#3}{ and~#2#1#3}
\Crefformat{appendix}{Appx.~\S#2#1#3}
\Crefformat{section}{~\S#2#1#3}
\crefformat{algorithm}{Alg.~#2#1#3}
\usepackage{xcolor}

\newcommand{\syn}{T2I\xspace}

\newcommand{\xmark}{\textcolor{red}{\ding{55}}}
\newcommand{\cmark}{\textcolor{blue}{\ding{51}}}

\newcommand{\graycell}{\cellcolor[gray]{.90}}

\newcommand{\usc}{$^\diamond$}
\newcommand{\harvard}{$^\dagger$}
\newcommand{\microsoft}{$^\varkappa$}

\iccvfinalcopy 

\def\iccvPaperID{5245} 

\ificcvfinal\fi

\begin{document}

\title{Beyond Generation: Harnessing Text to Image Models for \\ Object Detection and Segmentation
}
\author{
Yunhao Ge$^*$\usc~~~
Jiashu Xu$^*$\harvard~~~
Brian Nlong Zhao\usc~~~
Neel Joshi\microsoft~~~
Laurent Itti\usc~~~
Vibhav Vineet\microsoft\\
{\usc}University of Southern California~~~
{\harvard}Harvard University~~~
{\microsoft}Microsoft Research\\
{\small
\texttt{\{yunhaoge, briannlz, itti\}@usc.edu}~~~
\texttt{jxu1@g.harvard.edu}~~~
\texttt{\{neel, Vibhav.Vineet\}@microsoft.com}
}
\\
\small{* = Equal Contribution;~~~ \url{https://github.com/gyhandy/Text2Image-for-Detection}}
}
\maketitle
\ificcvfinal\thispagestyle{empty}\fi

\begin{abstract}

We propose a new paradigm to automatically generate training data with accurate labels at scale using the text-to-image synthesis frameworks (e.g., DALL\text{-}E, Stable Diffusion, etc.). 
The proposed approach\footnote{This is an extension of DALL-E for detection \cite{ge2022dalle}} decouples training data generation into foreground object generation, and contextually coherent background generation.
To generate foreground objects, we employ a straightforward textual template, incorporating the object class name as input prompts. This is fed into a text-to-image synthesis framework, producing various foreground images set against isolated backgrounds. A foreground-background segmentation algorithm is then used to generate foreground object masks. 
To generate context images, we begin by creating language descriptions of the context. This is achieved by applying an image captioning method to a small set of images representing the desired context. These textual descriptions are then transformed into a diverse array of context images via a text-to-image synthesis framework. Subsequently, we composite these with the foreground object masks produced in the initial step, utilizing a cut-and-paste method, to formulate the training data.
We demonstrate the advantages of our approach 
on five object detection and segmentation datasets, including Pascal VOC and COCO.
We found that detectors trained solely on synthetic data produced by our method achieve performance comparable to those trained on real data (Fig.~\ref{fig:teaser}). Moreover, a combination of real and synthetic data yields even much better results. Further analysis indicates that the synthetic data distribution complements the real data distribution effectively. Additionally, we emphasize the compositional nature of our data generation approach in out-of-distribution and zero-shot data generation scenarios. 
We open-source our code at {\url{https://github.com/gyhandy/Text2Image-for-Detection}.}
\end{abstract}

\section{Introduction}
\vspace{-5pt}

\label{sec:intro}

\begin{figure}[t]
\begin{center}
\includegraphics[width=\linewidth]{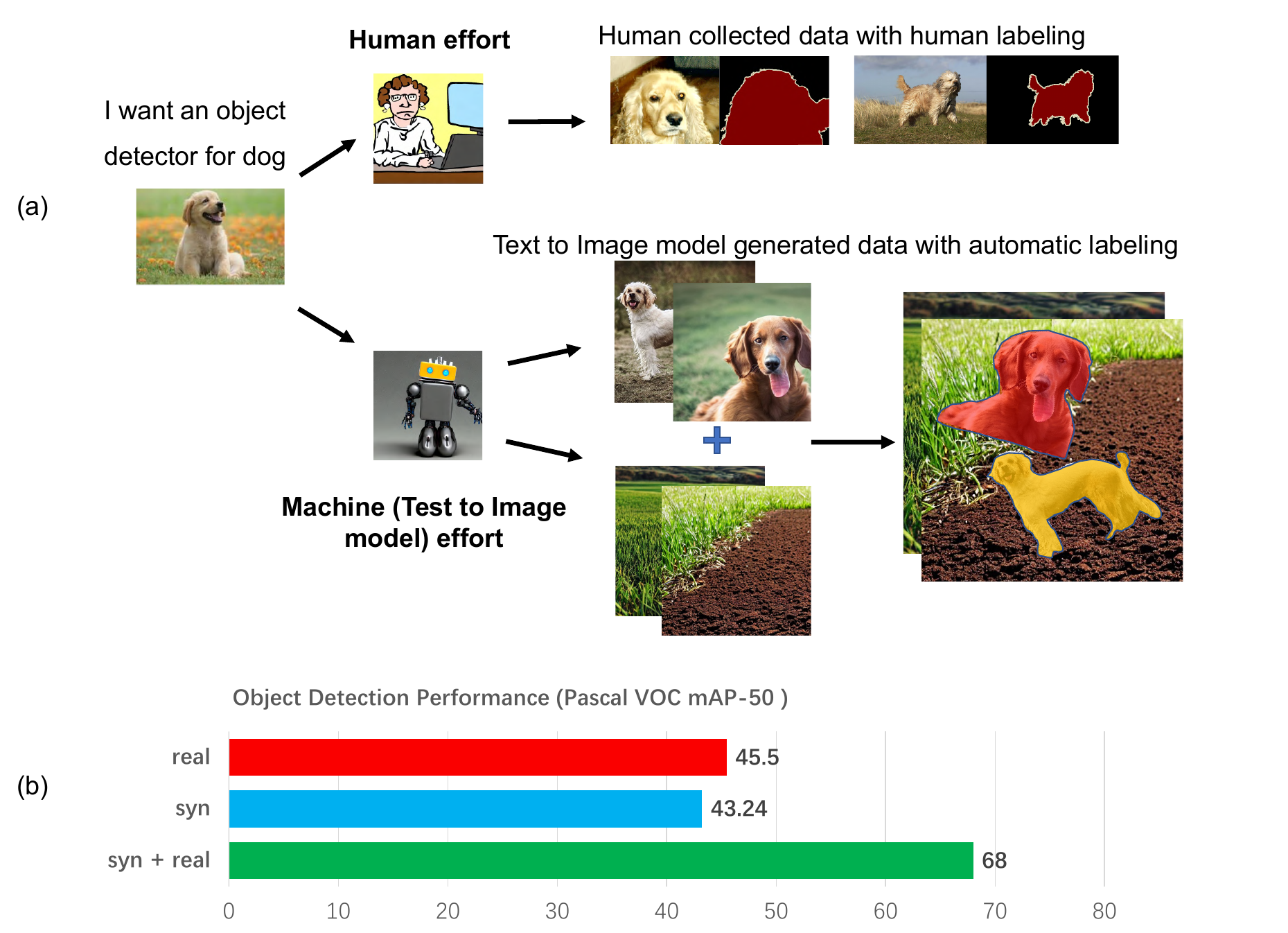}
\end{center}
   \caption{
   (a) Comparison of DALL-E for detection pipeline and traditional human-centric pipeline (b) Using pure synthetic data from the text-to-image model (\texttt{syn}) could lead on-par performance with using all real data (\texttt{real}), mixing real and synthetic (\texttt{syn + real}) gives strong performance gains (+22.5 mAP).
   }
\label{fig:teaser}
\end{figure}

Training modern deep learning models necessitates large labeled datasets \cite{ren2017faster,he2016deep,Shelhamer_2017}. Yet, acquiring such datasets poses significant challenges due to the high costs and extensive human effort involved, making the process both expensive and time-intensive.
This leads us to a pivotal question: \textit{Is it possible to efficiently generate large-scale labeled data that also delivers high accuracy for subsequent tasks?}
We believe that any such approach should satisfy these qualities (\Cref{table:advantage}): minimal human involvement, automatic generalization of the images for any new classes and environments, scalable, generation of high quality and diverse set of images, explainable, compositional, and privacy-preserving.

To this end, synthetic techniques could be used as promising avenues for generating labeled data for training computer vision models.
One popular approach is to use computer graphics to generate data \cite{richter2016playing,song2016ssc,FallingThings}. 
Despite their potential, these strategies predominantly hinge on acquiring 3D models of both objects and scenes. This often necessitates specialized skills, such as 3D modeling expertise, which inherently curtails the scalability of such methods.
Another viable approach is using NeRF based rendering \cite{mildenhall2020nerf,NeuralSim}.
These strategies typically entail retraining models to accommodate new object classes. As such, they lack the flexibility to effortlessly scale across a vast array of object classifications.
Lastly, a third approach is
object cut and paste \cite{dwibedi2017cut}, 
which pastes foregrounds onto the backgrounds.
However, such an approach demands a diverse and reliable source of foreground object masks as well as coherent backgrounds, which can be challenging to procure.

Recently, there has been a revolutionary advancement in text-to-image (\syn) synthesis models such as DALL-E \cite{ramesh2021zero}, Stable Diffusion \cite{rombach2022high}, RU-DALLE \cite{sberbank31:rudalle}, 
CogView \cite{ding2021cogview}, 
Imagen \cite{saharia2022photorealistic}, 
MUSE \cite{muse}, and eDiff-I \cite{eDiffI}.
Not only are these models capable of crafting high-quality images, but they also excel in illustrating intricate scenes, understanding semantics, and capturing the compositional nuances of the real world.
Therefore,
they might act as a natural bridge between humans and image synthesis. 
Nevertheless, despite their capability to function as synthetic data creators, these models lack the ability to produce region-level bounding boxes or pixel-level segmentation. Consequently, they remain unsuitable for downstream tasks such as object detection or instance segmentation.

\begin{table}
\begin{center}
\resizebox{0.48\textwidth}{!}{
\begin{tabular}{c|c c c c c c c}
\hline
    Method  & Quality & Less Human & Adapt & Scalable  & Explain. & Privacy & Comp. \\
\hline
  Human capture & \cmark & \xmark & \xmark & \xmark & \cmark & \xmark & \cmark \\
  Web image    & \cmark & \cmark & \xmark & \cmark & \xmark & \xmark  & \xmark \\
  Public dataset  & \cmark & \xmark & \xmark & \cmark & \xmark & \xmark  & \xmark \\
  Generative models   & \cmark & \cmark & \xmark & \cmark & \xmark & \cmark & \xmark \\
  Ours & \cmark & \cmark & \cmark & \cmark & \cmark & \cmark & \cmark \\
\hline
\end{tabular}}
\end{center}
\vspace{-10pt}
\caption{
Desired quality of context generation method: images should be high quality and diverse, less human involvement, generalization of the images for any new environment, scalable, explainable, privacy-preserving, and compositional.
}
\label{table:advantage}
\end{table}

In this work, we explore if \textit{these T2I models can be harnessed to generate expansive training datasets equipped with accurate labels, specifically tailored for tasks such as object detection and instance segmentation.}
Moreover, we primarily focus on \textit{low-resource regime} wherein downstream users have access to a limited training dataset but demand a high-performing, robust, and broadly applicable object detector or instance segmentation.
We opt for this context because acquiring on-demand human annotations tailored to the specific requirements of each individual downstream user can be prohibitively expensive, making the role of synthetic datasets even more crucial.

We propose a novel data generation paradigm that use \syn to produce large-scale high-quality and contextually-coherent pseudo-labeled datasets with minimal human involvement. 
Essentially, we retrieve pertinent information to curate a specific dataset aimed at enhancing 
downstream discriminative object detectors and instance segmentation, from the general knowledge base of generative \syn.

Our pipeline is bifurcated into two components: foreground object mask generation and contextual background generation (\Cref{fig:capgen}).
For zero-shot scenarios where only class names are available,
we employ a set of simple templates for each of the interested objects, such as ``A photo of dog in a white background'' to elicit dog-related images from the \syn.
Given that our templates are devised to position objects against easily-segmentable backdrops, straightforward background-foreground segmentation techniques can be employed to extract precise foreground object masks. 
For generating backgrounds, we employ analogous templates to craft clear backgrounds devoid of any target objects. 
After CLIP filtering, 
foreground object masks are pasted onto the backgrounds via cut and paste \cite{dwibedi2017cut}, resulting in the final pseudo-labeled dataset.
For few-shot scenarios where additional real training images are available, 
we specifically produce contextually-aligned coherent backgrounds from these shots as 
background context plays a pivotal role in learning strong object recognition models \cite{divvala2009empirical, dvornik2018modeling}.
For example, placing airplanes and boats in their natural context helped to improve accuracy, \eg airplanes are generally found in the sky and boats are on the water.
We caption each shot, extract contextual keywords (\eg ``grass field'' from ``A dog lying on grass field'') from these captions, contextually augment those captions (\eg ``A real photo of forest''), and produce coherent images by feeding \syn those augmented captions.

The proposed pipeline satisfies all desired properties of data generation (\Cref{table:advantage}).
It minimizes the requirement for human intervention in both the initial provision of training data and the subsequent steps of synthetic data generation.
\syn enables privacy-preserving and scalable generation.
We obtain explainable and compositional data generation by operating within the language domain before feeding to \syn.
Adding or removing objects or settings can be easily done. For example, a description as ``an environment with a table'' can be easily modified to a kitchen environment by utilizing the compositional properties of language as ``a kitchen environment with a table.''
Furthermore, even when out-of-distribution training instances are given, by simple language edit, we can easily force generated synthetic dataset to be better matched with the test distribution (\eg by substituting ``cartoon kitchen'' with  ``real kitchen'' to bridge the gap between cartoon training set and real test set).

Our main contributions are three-fold:
(1) We propose a language-driven compositional synthetic dataset generation that automatically generate large-scale high-quality foreground objects and contextually coherent backgrounds.
(2) We demonstrate strong performance gains across three downstream tasks and five widely-used benchmarks, under various low-resource regmines including 0, 1, and 10-shot settings.
(3) We highlight that the compositionality in our synthetic data generation results in a closer alignment with the test distribution, even when dealing with out-of-distribution training instances.
To the best of our knowledge, this is the first work to use vision and language models for generating object detection and segmentation datasets.
\section{Related works}
\label{sec:related_works}

\paragraph{Text-to-Image Synthesis Models.} 
%
T2I approaches like DALL-E \cite{ramesh2021zero}, RU-DALLE \cite{sberbank31:rudalle}, Stable Diffusion \cite{rombach2022high}, CogView \cite{ding2021cogview} have revolutionized high-quality image generation.
These approaches leverage benefits of using large transformer models trained using large vision and text data. Though they can generate high quality images of real world scenes, they are not able to generate ground truth labels for objects. For example, they are not able to provide bounding box and per-pixel annotation for objects. In our work, we propose an automatic approach to generate high quality images with ground truth bounding box and per-pixel labels from T2I models.
\vspace{-5mm}
\paragraph{Synthetic Data Generation.} 
A series of works on using synthetic data for training computer vision problems have been proposed. Some of them include using graphics pipeline or computer games to generate high quality labelled data \cite{richter2016playing,RichterHK_iccv17,RosSMVL_cvpr16,hodan2019photorealistic,TremblayTB_corr18,HandaPBSC_corr15a}. Generally using graphics pipeline requires having 3D models of both objects and environment, that may limit their scalability. Some of them use generative models (e.g., GAN) \cite{bowles2018gan,ge2020pose} or zero-shot synthesis \cite{ge2020zero} to augment datasets and remove bias. However, they need a relatively large initial dataset to train the model, and not easy to generalize to new domains.

The idea of pasting foreground objects on background images has emerged as a easy and scalable approach for large scale data generation. 
%
%
The idea has been used to solve vision problems like object instance segmentation tasks \cite{dwibedi2017cut}, object detection and pose estimation \cite{SuQLG15,hinterstoisser2017pre,rad2017bb8,TekinSF_corr17,tripathi2019learning}, and more \cite{Dosovitskiy_2015_ICCV, yun2021cut, ghiasi2021simple}.
%
%
These approaches generally require accurate foreground object masks. This limits their scalability. While we utilize a cut-and-paste approach, in contrast to previous works in this space, our work can generate foreground object masks for any new object class.

\vspace{-5mm}
\paragraph{Language for object recognition.} 
%
Vision-language based models have been developed for image captioning \cite{vinyals2015show,rennie2017self,anderson2018bottom,gurari2020captioning}, visual question answering tasks \cite{antol2015vqa,vedantam2019probabilistic,agrawal2018don,das2018embodied} and others  \cite{lu202012,yang2021causal,li2020unicoder}. In recent years, vision-language based models have been developed for self-supervised training. 
The recent CLIP appraoch \cite{radford2021learning} showed how training a model on large image-text pairs can generalize to several image classification datasets where current image based models performed very poorly. Such vision-language models havec been used to solve other tasks like object detection \cite{gu2021zero,kamath2021mdetr,li2022grounded} and semantic  segmentation \cite{li2022language}. These works demonstrate benefits of using language in solving vision tasks. However, these methods can not be used generate new data for any new tasks or environment settings. 
Large T2I synthesis models can be used to generate new data for a new task. 
%
A recent concurrent work X-Paste \cite{zhao2022x} has used Stable Diffusion to solve object detection. 
We are motivated by generation quality of these text to image generation methods in this work.

%


\section{Method}
\label{sec:method}

\begin{figure*}
\begin{center}
\includegraphics[width=\linewidth]{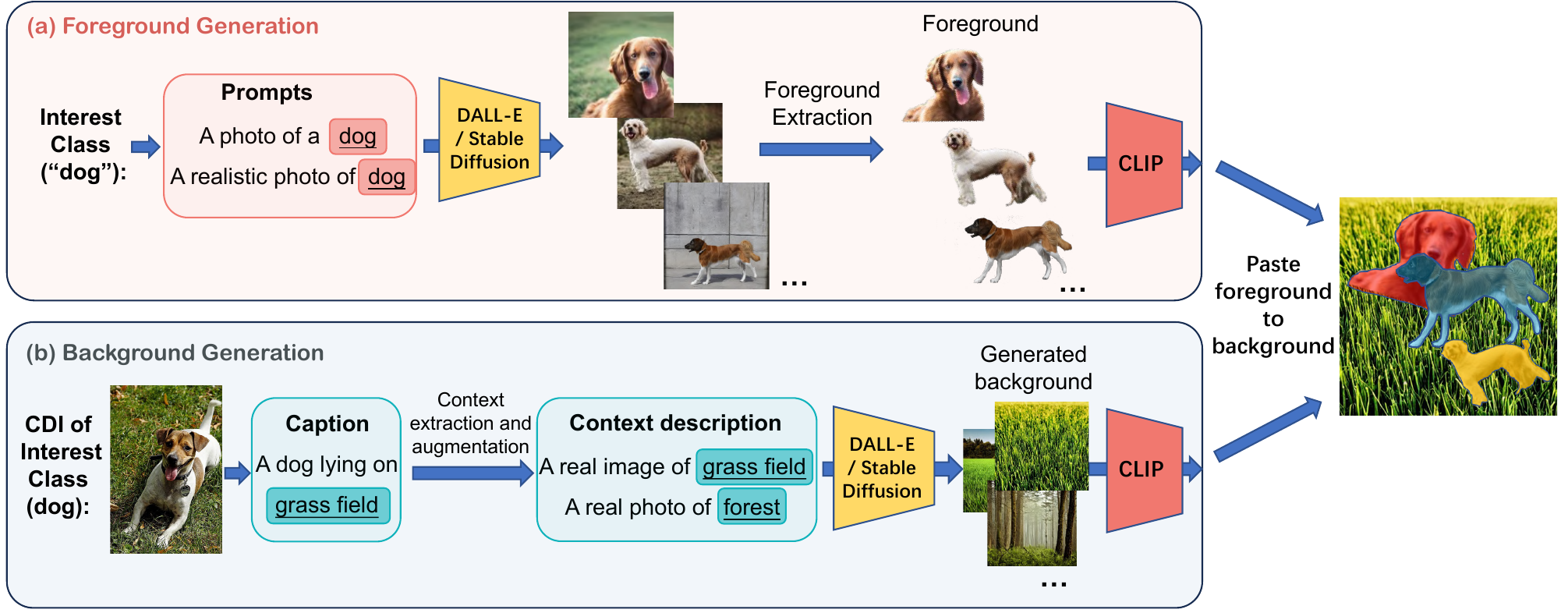}
\end{center}
    \caption{
    (a) \textbf{Foreground generation:} (top row, \Cref{sub:fg_gen}) verbalizes class name into templates understandable by \syn models \cite{ramesh2021zero, rombach2022high}, which synthesize desired foreground images with easy-to-separate backgrounds. Off-the-shelf foreground/background segmentation methods are then used to extract foreground segments from foreground images.
    (b) \textbf{Background generation:} (bottom row, \Cref{sub:bg_gen}) 
    an image captioning method (\eg SCST \cite{rennie2017self}) captions user-provided images (CDIs).
    Context words (\eg ``grass field'') are extracted and the augmented caption is feed into \syn to generate background images.
    (c) CLIP \cite{radford2021learning} is used (\Cref{sub:clip sanity check}) to maintain the quality of both foregrounds and backgrounds, as well as ensure that the generated images do not have unwanted class. 
    (d) Finally, we composite (\Cref{sub:cut paste}) the foreground segments and background images to obtain synthetic images with accurate labels.
}
\label{fig:capgen}
\end{figure*}

\begin{figure}[h]
\begin{center}
\includegraphics[width=\linewidth]{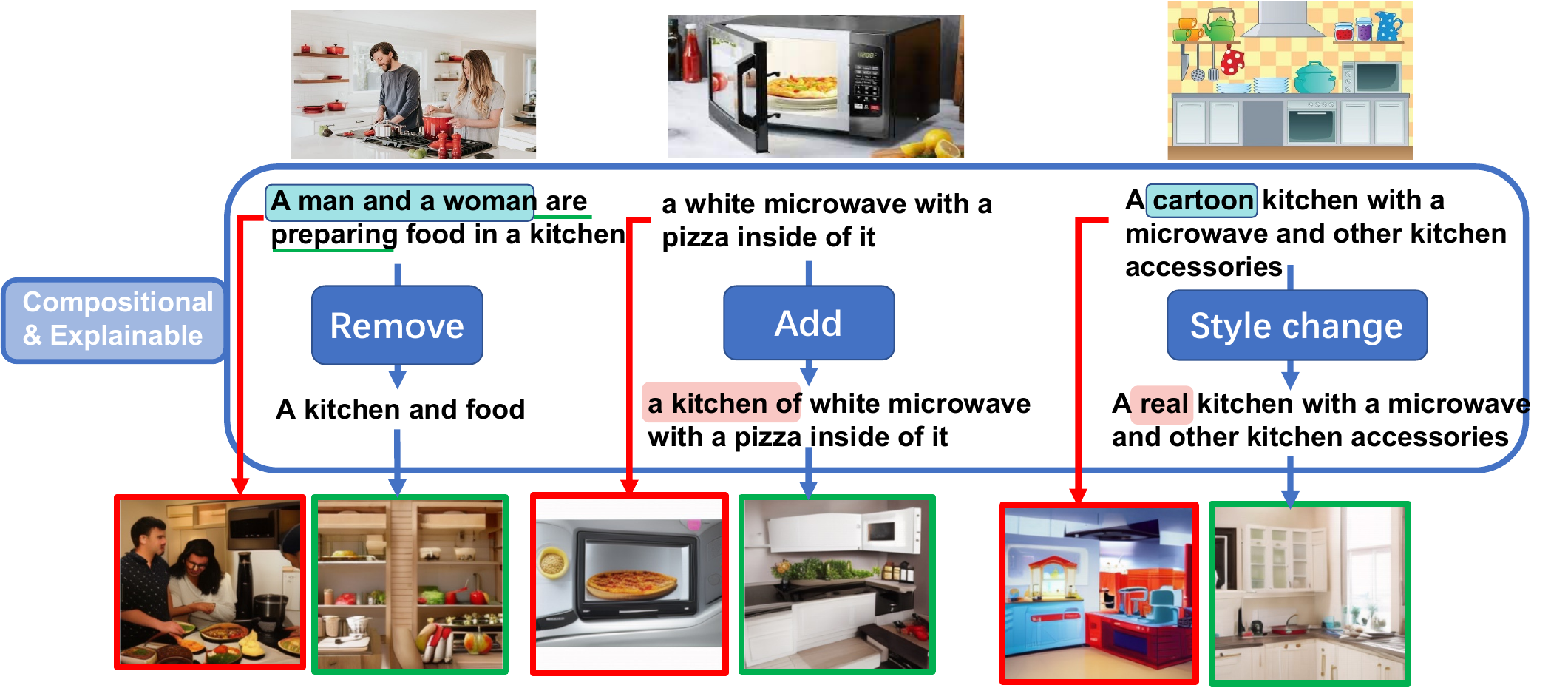}
\end{center}
\vspace{-5mm}
   \caption{
   When CDIs can not perfectly describe the real test scenario, the compositional property of language can help to correct the context description. 
   For instance, if the initial description contains noisy information ``man and a woman'', we can remove the noisy information to generate a congruent context description.
   Images with a \textcolor{red}{red} frame show the generated image without language intervention and the \textcolor{green}{green} frame shows the images after the intervention.
cccccbkrkkdllkdhegjbknulebflteevrrhtdjuefjeh
}
\label{fig:compositionality}
\end{figure}

We layout our method in \Cref{fig:capgen}.
This work aims to enhance object detection and instance segmentation models by efficiently generating a diverse and extensive collection of pseudo-labeled synthetic data using text-to-image models.
One key observation is that each image can be divided into backgrounds and foregrounds.
In this work, we propose to generate synthetic foreground objects (with mask
extraction) (\Cref{sub:fg_gen}) and contextually coherent backgrounds (\Cref{sub:bg_gen}) separately.
Subsequently, after CLIP filtering on both (\Cref{sub:clip sanity check}), the synthesized foregrounds are then composited onto the synthesized backgrounds via cut-paste \cite{dwibedi2017cut} (\Cref{sub:cut paste}).
Examples of synthetic datasets can be viewed in \Cref{fig:analysis}, \Cref{fig:final-composition-voc-smaller} and \Cref{fig:final-composition-voc}.

We utilize off-the-shelf text-to-image
generative models (\syn) \cite{ho2020denoising, rombach2022high, ho2022classifier} to generate both the foreground masks and contextual backgrounds.
Firstly, \syn efficiently compresses web-scale image-text data, ensuring both portability and scalability. The text-to-image generation model can produce an endless array of high-quality images, with our method offering a controllable generation process. Our experiments and subsequent analysis confirm that the synthetic data distribution effectively complements the real data distribution, as detailed in \Cref{sec:analysis}.
Also, the generative model can create novel scenarios that were not present in the training data (\Cref{sec:generalization}, \Cref{sec:compositionality}).
Furthermore, language-based generation facilitates compositionality (\Cref{sec:compositionality}).
Lastly, the synthetic nature of the data generation procedure ensures privacy preservation.

We underscore that our method, thanks to the capabilities of \syn in generating diverse images, particularly excels in low-resource regimes where ground truth data is limited.
Following the terminology of \cite{radford2021learning, li2023your},
in this work, we focus mainly on \textbf{zero-shot} (where no ground-truth data is provided but a list of desired objects) and \textbf{few-shot} (where one or ten images per class are given).
We also included experiments to show our methods improve using full-set as well (\Cref{tab:voc_object_detection}).

\begin{table*}[h]
    \centering
    \begin{tabular}{ccc}
    \hline
        A photo of <object> & A realistic photo of <object> & A photo of <object> in pure background \\
        \hline
        <object> in a white background &  <object> without background & <object> isolated on white background \\
        \hline
    \end{tabular}
    \caption{Six manually designed templates for generating foreground images zero-shot. Here <object> will be replaced by label names such as ``bus''. The design philosophy is to put objects in a clean background for easy foreground extraction.}
    \label{tab:fg_template}
\end{table*}

\begin{table*}[h]
    \centering
    \begin{tabular}{cccc}
    \hline
    empty living room & empty kitch & blue sky & empty city street, color \\
    \hline
    empty city road, color & empty lake & empty sea & railway without train \\
    \hline
    empty railway, color & trees & forest & empty street, colored \\
    \hline
    farms & nature & empty farm & stable \\
    \hline
    \end{tabular}
    \caption{Sixteen handcraft templates for generating coherent background images zero-shot. The full template is ``A real photo of <context>'' where <context> is substituted with one of the above 16 places. The design philosophy is to create natural images but without any interested objects (thus ``empty'') since we would not have segmentation labels for those objects if they are generated.}
    \label{tab:bg_template}
\end{table*}

\subsection{Zero-shot Foreground Synthesis}
\label{sub:fg_gen}

In object detection or segmentation, the quality of foreground objects significantly influences the performance of downstream tasks
\cite{ghiasi2021simple}.
This poses a significant challenge in low-resource scenarios, as training a model that can effectively generalize with limited exposure to seen foregrounds is inherently difficult. v
To address this, we aim to harness the available prior knowledge to generate a vast and diverse collection of high-quality foregrounds\footnote{Samples of the synthesized images can be found in \Cref{fig:capgen} and \Cref{fig:analysis}.}.

\begin{figure*}[h]
\begin{center}
\includegraphics[width=0.95\linewidth]{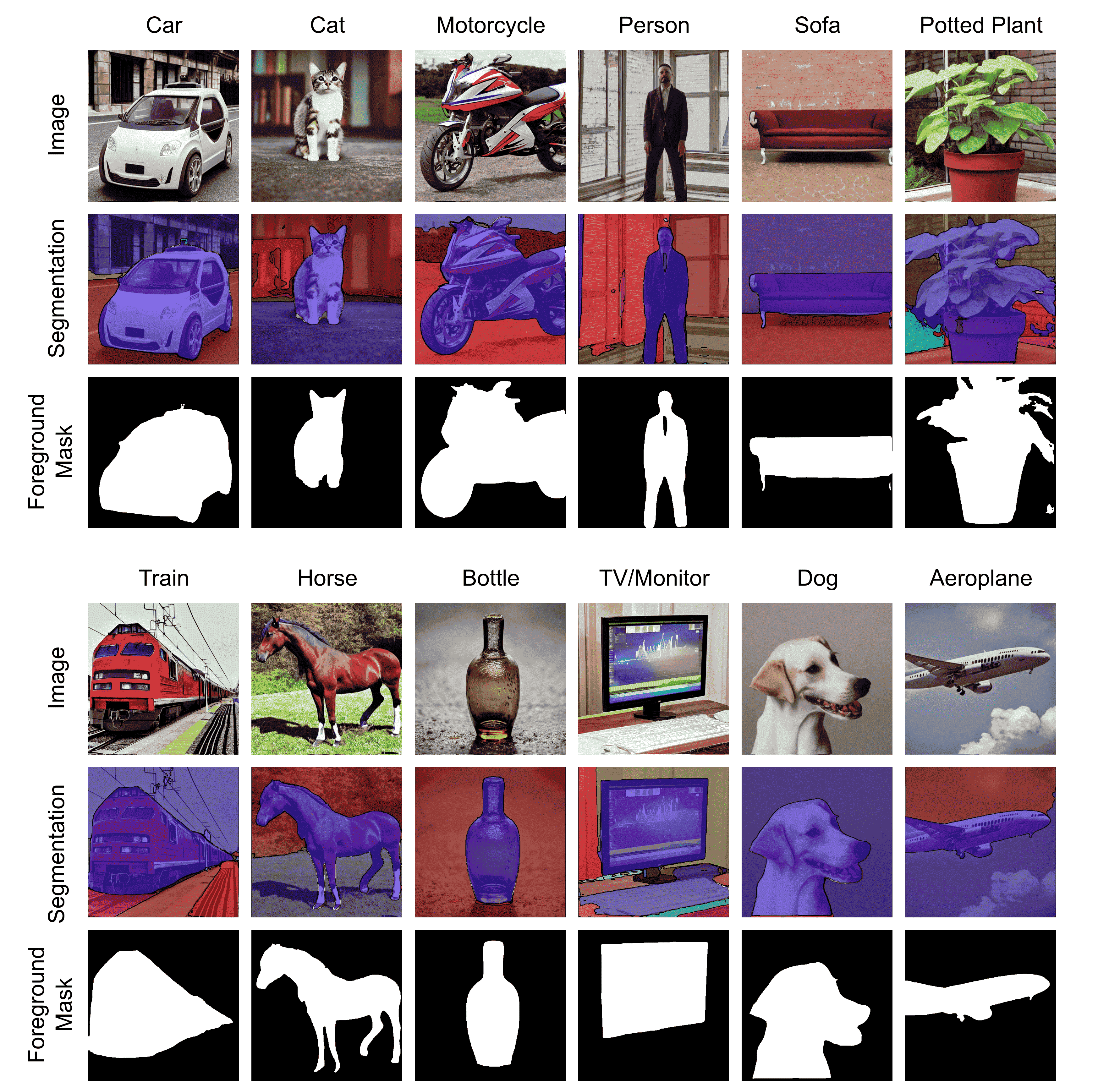}
\end{center}
   \caption{Examples of foreground segment extraction from generated images. 
   }
\label{fig:fg-extract}
\end{figure*}

Specifically, we manually design six fixed prompt templates
(\Cref{tab:fg_template}), where <object> is substituted with the class name like ``dog.''
We then feed those verbalized templates into the \syn, which generates high-quality iconic object images.
The templates are designed to elicit images where interested objects are centered on a simple isolated background, enabling straightforward extraction of the foreground object masks
using an unsupervised segmentation method \cite{qi2021open}, as analyzed in \Cref{sec:analysis}. Examples of generated foregrounds and extracted masks can be viewed in \Cref{fig:analysis} and \Cref{fig:fg-extract}.

\subsection{Language-driven Context Synthesis} \label{sub:bg_gen}

Merely having high-quality foregrounds is insufficient for training a robust downstream model. It is equally crucial to position those foregrounds within suitable in-context backgrounds \cite{divvala2009empirical, mottaghi2014role, dvornik2018modeling, zhang2020learning, lee2018context, yun2021cut}.
In our experiments (\Cref{sec:ablation}) we discovered that the choice of backgrounds can significantly impact performance, with different backgrounds having the potential to degrade results dramatically. Thus, we propose to utilize \syn to generate coherent backgrounds as relevant ``context'' for downstream models.

Other than creating contextually coherent backgrounds, one additional benefit is that, as contexts are described in natural language, compositional generation is possibly by editing the descriptions \eg adding or removing a keyword.
For example, in \Cref{fig:compositionality} the word {``kitchen''} can be added or {``people''} can be removed to align more closely with the test distribution.

\subsubsection{Zero-shot scenario}
\label{sub:bg_gen:zero-shot}
Similar to \Cref{sub:fg_gen}, for the zero-shot scenario, we design sixteen background templates
(\Cref{tab:bg_template})
intended to generate backgrounds without interested objects, avoiding
``misplaced'' objects with unknown object masks.
The selection of background templates is primarily based on a cursory examination of various images from the training set, Pascal VOC \cite{everingham2010pascal} and MS COCO \cite{lin2014microsoft}, in our case.
However, it is important to highlight that this process is minimal, and only high-level background descriptions are required.

\subsubsection{Few-shot scenario}
\label{sub:bg_gen:few-shot}
In a more relaxed few-shot setting, a few exemplars, dubbed \emph{Context Description Images (CDI)}, are given, all sampled from a distribution that is contextually close to the test distribution.
CDIs provide valuable contextual cues.
We term this setting ``few-shot'' but note that provided CDIs \textbf{do not have to be sampled from the training distribution} (\Cref{fig:compositionality}). \Eg
if the test distribution includes a kitchen environment, the small set of kitchen images can be taken from any public dataset or web images (\Cref{sec:generalization}, \Cref{sec:compositionality}).
Our goal is to mine beneficial context to encourage the synthesized background distribution to more closely match the test distribution by imposing the context constraint.
Lastly, we note that our method significantly reduces human labor, since our method requires a minimal number of CDI.
In fact, our method works sufficiently well using \textit{as little as 1 CDI} (\Cref{fig:evidence for model able to generate a lot from single CDI }, \Cref{sec:compositionality}).

\noindent{\textbf{Context Description.}}
We first describe the context information in natural language via image captioning.
We leverage self-critique sequence training (SCST) \cite{rennie2017self} to generate a set of diverse textual captions for input CDIs.
Yet we note that our method is agnostic to the image captioning method and any other methods \eg \cite{anderson2018bottom, gurari2020captioning, li2022blip} can also be applied.

\noindent{\textbf{Context Extraction and Augmentation.}}
Our focus is primarily on the background rather than the objects themselves, as the generated objects do not have labels.
Therefore we need to extract background contexts out of the captions that might contain objects.
We built a simple parser that extracts context words (\eg, ``grass field'') or removes unwanted objects (\eg detecting nouns such as ``dogs'').
These cleansed captions are then transformed into a few augmented contexts (\eg ``A real photo of grass field'' and ``A real photo of forest'') via simple heuristics.
While it is possible to automate the entire context extraction process using large language models \cite{brown2020language, openai2023gpt4}, we leave that as future work.

\noindent{\textbf{Context-guided Synthesis.}}
We feed augmented contexts into the \syn to produce a diverse set of contextually coherent backgrounds.
As these augmented contexts are derived from the CDIs, they are more contextual-aligned with the test distribution; also they should contain no objects since these are removed in the extraction process.

\subsection{CLIP Sanity Check}
\label{sub:clip sanity check}
Due to the limitations of \syn, we observed that it occasionally generated irrelevant or nonsensical images that strongly deviated from the input text.
Moreover, 
\syn method learns associations between the context and objects,
leading to instances where horses would frequently appear in the context of a ``stable,'' even when the text explicitly states ``An empty stable.''
Thus, in \Cref{sec:ablation} we find it
indispensable to post-process generated foregrounds and backgrounds via CLIP \cite{radford2021learning} filtering.
Specifically, we use CLIP to rank images via two rules: images are semantically faithful to the input text and semantically dissimilar to any interest classes.
This step ensures the generations align more closely with desired semantics.

\subsection{Cut-paste Composition} \label{sub:cut paste}
To create the final pseudo-labeled training data, we composite foreground object masks (\Cref{sub:fg_gen}) onto the backgrounds (\Cref{sub:bg_gen}) using cut-paste \cite{dwibedi2017cut}.
At each step, a set of four foreground object masks is selected and pasted into a sampled background image, and such procedure is repeated until all foreground masks are pasted.
The foreground mask, after random 2D augmentation such as rotation and scaling, is pasted on a random location in the image.
In addition, following \cite{dwibedi2017cut, ghiasi2021simple}, we apply a Gaussian blur on the object boundary with blur kernel $\sigma$ to blend pasted foregrounds.
\section{Experiments}
\label{sec:experiment}

\begin{table*}[htpb]
    \centering
    \small
    \begin{tabular}{c|lll|cc}
    \toprule[1pt]
        \textbf{\#CDI} & \textbf{Method} & \textbf{Foreground} & \textbf{Background} & \textbf{mAP@50} & \textbf{mAP} \\ \hline
        1,464 (Fullset) & Pure Real \cite{he2017mask} & Real & Real & 45.50 & 17.00 \\
        \hline
        0 (0 shot) &\graycell Pure Syn &\graycell Syn &\graycell Syn &\graycell 43.24 &\graycell 19.78 \\
        \hline
        \multirow{6}{*}{\rotatebox[origin=c]{0}{$20 \cdot 1$ (1 shot)}} & Pure Real \cite{he2017mask}  & Real & Real & 0.14 & 0.04 \\
        & \ \ + cut paste \cite{dwibedi2017cut} & Real & Real  & 6.03 & 2.07 \\
        &\graycell  Syn Fg  &\graycell Syn + Real &\graycell Real &\graycell 37.97 &\graycell 17.53 \\
        &\graycell  Pure Syn  &\graycell Syn &\graycell Syn &\graycell 44.24 &\graycell 20.63 \\
        &\graycell  Syn + real  &\graycell Syn + Real &\graycell Syn + Real &\graycell \textbf{45.62} \textcolor{red}{\small (+39.59)} &\graycell \textbf{21.45}  \textcolor{red}{\small (+19.38)} \\
        \hline
        \multirow{7}{*}{\rotatebox[origin=c]{0}{$20 \cdot 10$ (10 shot)}} & Pure Real \cite{he2017mask}  & Real & Real & 9.12 & 2.35 \\
        & \ \ + cut paste \cite{dwibedi2017cut} & Real & Real & 29.60 & 10.82 \\
        & \graycell Syn Fg & \graycell Syn + Real & \graycell Real &\graycell 48.14 &\graycell 21.62 \\
        & \graycell Pure Syn & \graycell Syn &\graycell Syn &\graycell 45.12 &\graycell 22.38 \\
        & \graycell Syn + real &\graycell  Syn + Real &\graycell Syn + Real &\graycell \textbf{51.82}  \textcolor{red}{\small (+22.22)} &\graycell \textbf{25.87}  \textcolor{red}{\small (+15.05)} \\ \hhline{~-----}
        & \graycell Syn + 1,464 real &\graycell Syn + 1,464 Real &\graycell Syn + 1,464 Real &\graycell \textbf{68.38} \textcolor{red}{\small (+22.88)} &\graycell \textbf{35.96} \textcolor{red}{\small (+18.96)} \\
        \bottomrule[1pt]
    \end{tabular}
    \caption{
    We harness \syn (Stable Diffusion) to generate a large-scaled high-quality synthetic foregrounds and backgrounds, and improve VOC object detection.
    Column mAP is computed as the average of IoU ranging from 50 to 95 with step size 5.}
    \label{tab:voc_object_detection}
\end{table*}

We now present experiments to demonstrate the effectiveness of the large-scale synthetic data generated using our proposed approach.
In \Cref{sec:det}, we first provide detailed results on object detection task on the Pascal VOC \cite{everingham2010pascal} and COCO \cite{lin2014microsoft} datasets in low-resource regimes including zero-shot, 1-shot and 10-shot settings.
We are interested in low-resource settings, which are more common in practice yet highly challenging due to the limited information provided. In this case, CDIs are the shots given.
We emphasize that our method particularly shines in such settings due to the capability to create a large diverse set of high-quality coherent synthetic training datasets.
In addition, we present ablation studies (\Cref{sec:ablation}) to investigate the impact of different design choices. Importantly, further analysis (\Cref{sec:analysis}) demonstrates that the synthetic data distribution effectively complements the real data distribution.

Next, in \Cref{sec:generalization}, we show that our method can generalize to more tasks, including instance segmentation tasks on VOC and COCO, and instance detection tasks on GMU-Kitchen \cite{georgakis2016multiview}, Active Vision \cite{ammirato2017dataset}, and YCB-video datasets \cite{xiang2017posecnn}.
Finally, in \Cref{sec:compositionality}, we also provide results highlighting the compositional nature of our data generation process.

\noindent{\textbf{Model, training, and evaluation criterion.}} We use MaskRCNN \cite{ren2017faster} with a ResNet-50 \cite{he2016deep} backbone for compatibility among object detection, instance segmentation, and object instance detection tasks. We set the learning rate as $0.001$ with a weight decay $0.0005$ and train the models to convergence for both baselines and our approaches.
We report mean average precision (mAP) for the results.
In \Cref{sec:ablation} we additionally experiment with MaskRCNN with DINO-pretrained ResNet-50 \cite{caron2021emerging} as well as the recent transformer-based method EVA \cite{fang2023eva}.

\noindent{\textbf{Variants of synthetic dataset.}}
We evaluate three variants of our methods:
\begin{itemize}[leftmargin=*, topsep=1pt, wide=\parindent]
\itemsep-0.2em
\item \textbf{Pure Syn} uses purely synthetic foregrounds (\Cref{sub:fg_gen}) and zero-shot backgrounds (\Cref{sub:bg_gen:zero-shot}) for zero-shot.
    For few-shot settings where CDIs are available, contextual backgrounds from CDIs (\Cref{sub:bg_gen:few-shot}) are used on top of template backgrounds.
\item \textbf{Syn Fg} pastes synthetic foregrounds (\Cref{sub:fg_gen}) on real backgrounds. Note that foregrounds from original real backgrounds are retained.
\item \textbf{Syn + Real} further incorporates synthetic backgrounds.
    In other words, synthetic datasets are blended with the entire real dataset.
\end{itemize}
We mainly compare our methods with \textbf{Pure Real}, which trains MaskRCNN fully-supervised using the available real training set, and \textbf{Pure Real + cut paste}, which utilizes cut-paste \cite{dwibedi2017cut, ghiasi2021simple} to generate a relevant synthetic dataset from real images.
We note that the latter is an upper bound of what can be achieved with provided real images without leveraging external sources, such as \syn, as in our work.

\subsection{Object Detection}
\label{sec:det}
\begin{table*}[htpb]
    \centering
    \small
    \begin{tabular}{c|lll|cc}
    \toprule[1pt]
        \textbf{\#CDI} & \textbf{Method} & \textbf{Foreground} & \textbf{Background} & \textbf{mAP@50} & \textbf{mAP} \\ \hline
        0 (0 shot) &\graycell Pure Syn &\graycell Syn &\graycell Syn &\graycell 16.30 &\graycell 8.40 \\
        \hline
        \multirow{5}{*}{\rotatebox[origin=c]{0}{$80 \cdot 1$ (1 shot)}} & Pure Real \cite{he2017mask}  & Real & Real & 1.47 & 0.92 \\
        & \ \ + cut paste \cite{dwibedi2017cut} & Real & Real  & 2.89 & 1.23 \\
        &\graycell  Syn Fg  &\graycell Syn + Real &\graycell Real &\graycell 17.87 &\graycell 8.64 \\
        &\graycell  Pure Syn  &\graycell Syn &\graycell Syn &\graycell 16.80 &\graycell 8.59 \\
        &\graycell  Syn + real  &\graycell Syn + Real &\graycell Syn + Real &\graycell \textbf{20.82} \textcolor{red}{\small (+17.93)} &\graycell \textbf{10.63}  \textcolor{red}{\small (+9.40)}  \\
        \bottomrule[1pt]
    \end{tabular}
    \caption{Stable Diffusion generated foregrounds and contextual backgrounds enhance object detection on COCO dataset.}
    \label{tab:coco_object_detection}
\end{table*}

For object detection, we present zero-/one-/ten-shot results on PASCAL VOC 2012 \cite{everingham2010pascal} in \Cref{tab:voc_object_detection} and zero-/one-shot results on MS COCO \cite{lin2014microsoft} in \Cref{tab:coco_object_detection}. VOC encompasses 20 target objects, with 1,464 images for training and 1,449 for validation. While there exists an augmented training set comprising 10,582 images, it lacks instance segmentation masks, which are required by cut-and-paste augmentation. Given our emphasis on low-resource scenarios, we prioritize smaller training sets. MS COCO features 80 target objects with a total of 120k training instances.

\noindent{\textbf{Implementation details of synthetic datasets.}}
We use the instance segmentation mask labels from the training set as our real foreground masks, and CDI is the number of training images in this case.
Take 0-/10-shot VOC as example,\footnote{1-shot VOC and MS COCO is similar and omit for briefity.} detailed statistics is provided in \Cref{tab:details_for_voc_syn_dataset}.
For all three variants of synthetic datasets, we leverage
fixed foreground templates to generate high-quality foregrounds (\Cref{tab:fg_template}, \Cref{sub:fg_gen}).
For each of the 20 objects, we generate 500 images for each of the six templates, then use CLIP (\Cref{sub:clip sanity check}) to downselect 200 best images, thus aggregating to a total of 24k synthetic foregrounds.

In the \textbf{0 shot Pure Syn} scenario, we lack access to any training instances and must depend on fixed background templates to generate coherent backgrounds, as detailed in \Cref{tab:bg_template} and \Cref{sub:bg_gen:zero-shot}. For each of the 16 templates, we produce 600 images. Notably, the generated images are of high caliber and typically exclude target objects. As a result, we filter out 5\% of the images using only CLIP, as described in \Cref{sub:clip sanity check}, leaving us with 9,120 synthetic backgrounds. Onto each of these backgrounds, we superimpose four synthetic foregrounds, as outlined in \Cref{sub:cut paste}. This results in a synthetic dataset complete with segmentation labels for the foregrounds. We repeat this procedure until the count of final pseudo-labeled images reaches 60k.

In the \textbf{10 shot Pure Syn}, we have an additional 200 CDIs at our disposal. 
Instead of directly training on these real images, they serve to facilitate the creation of more contextually coherent backgrounds through context-guided synthesis, as detailed in \Cref{sub:bg_gen:few-shot}. 
Specifically, for each CDI, two captions are generated, from which 80 images are produced. Using CLIP, we then narrow this down to 30 images per CDI, as described in \Cref{sub:clip sanity check}. 
This results in a total of $2 \times 200 \times 30 = 12\text{k}$ contextually relevant backgrounds.
We then apply the cut-paste method, as outlined in \Cref{sub:cut paste}, using the \textit{same} zero-shot synthetic foregrounds on these expanded backgrounds, culminating in a final dataset of 60k.

For the \textbf{10 shot Syn+Real},  our training encompasses not just the synthetic datasets from Pure Syn, but also the actual 200 images. Within those $20 \times 10$ real images, we also obtain multiple real foregrounds: in our selection of 10 shot data, there are 541 unique foregrounds.

\begin{table*}[h]
\begin{center}
\small
\begin{tabular}{l|c|c|c|c}
\hline
Experiment & \# Real images & \# Foreground & \# Background & \# Training set size\\
\hline
Fullset (Pure Real) & 1464 & - & - & 1,464\\
0-shot (Pure Syn) & 0 & 24k & 9120 & 60k \\
10-shot (Pure Real) & 200 & - & -  & 200 \\
10-shot (Pure Real + cut paste) & 200 & 200 & 200 & 60k\\
10-shot (Syn Fg) & 200 & 24k & 200 & 60k\\
10-shot (Pure Syn) & 200 & 24k & 9120+12k & 60k\\
10-shot (Syn + Real) & 200 & 24k+541 & 9120+12k+200 &  60k\\
10-shot (Syn + Real) + 1464 & 1464 & 24k+541 & 9120+12k+200 & 60k + 1464\\
\hline
\end{tabular}
\end{center}
\caption{Detail statsitics of synthetic datasets created for VOC.}
\label{tab:details_for_voc_syn_dataset}
\end{table*}

\begin{figure}[b]
    \centering
    \includegraphics[width=\linewidth]{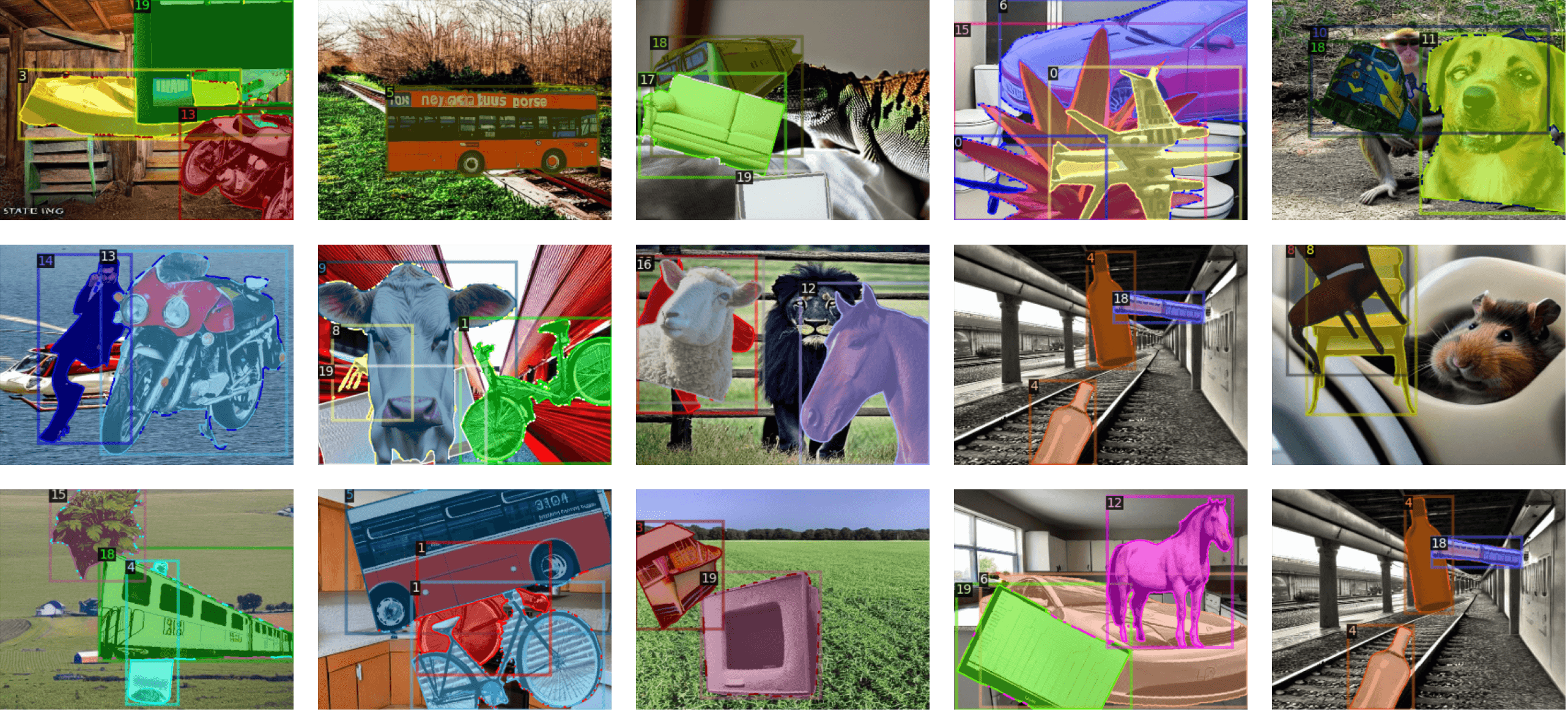}
    \caption{
    Pseudo-labeled synthetic images generated by our pipeline.}
    \label{fig:final-composition-voc-smaller}
\end{figure}

\begin{figure*}[h]
\begin{center}
\includegraphics[width=\linewidth]{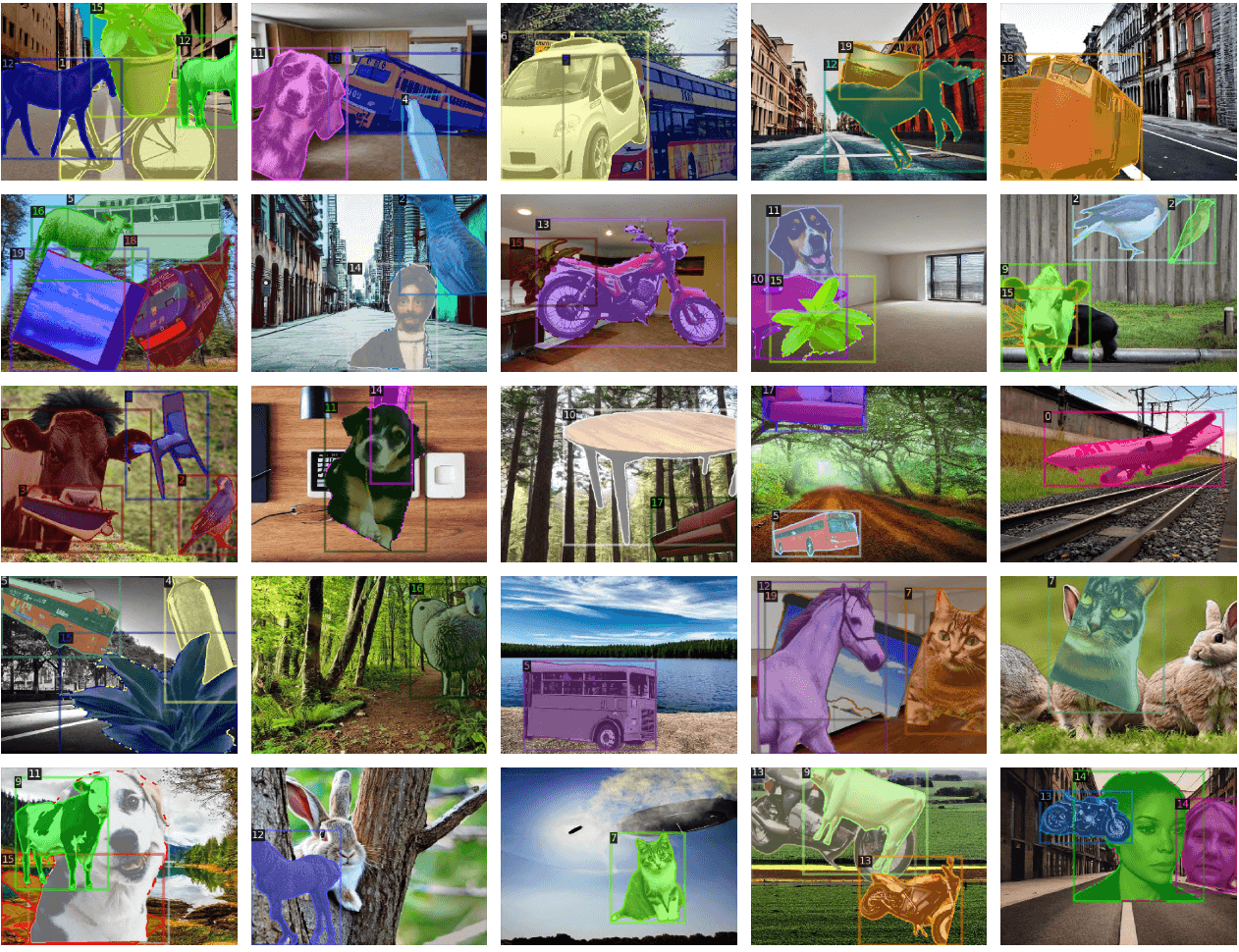}
\end{center}
   \caption{Training images generated by our pipeline for PASCAL VOC dataset: both foreground object and background context images are generated by our method with Stable Diffusion. }
\label{fig:final-composition-voc}
\end{figure*}

\Cref{fig:final-composition-voc-smaller} and \Cref{fig:final-composition-voc} show more example training images generated by our pipeline on PASCAL VOC dataset: both foreground object and background context images are generated by our method with Stable Diffusion.

We use Stable Diffusion \cite{rombach2022high} as the \syn, which takes 20 P40 GPUs 50h to generate 10-shot synthetic dataset.
We emphasize that
generation is a
one-time process that can train various downstream tasks and
models, and that 
our method is automatic and requires little human involvement to collect synthetic data.
Collecting COCO or VOC
manually takes much longer time with extra privacy or scalability issues.

\noindent{\textbf{Synthetic foregrounds and backgrounds benefit downstream task.}}
On VOC, we first notice that
a model trained solely on synthetic data in the absence of \emph{any} real images (0 shot Pure Syn) achieves comparable performance to a model trained on 1.4k real images (Pure Real).
This suggests that synthetic datasets can effectively improve downstream performance when only prior knowledge of interested objects is known.
We then observe that Pure Syn performance improves as adding CDIs, reinforcing our assumption that contextual information encoded in the backgrounds provides valuable cues in learning.
Further, we note that Syn Fg in few-shot setting outperforms Pure Real + cut paste significantly, implying that
inclusion of large-scale synthesized foregrounds enables the detection model to learn a more comprehensive understanding of objects due to the diversity and coherency of the synthetic data.
Lastly, further performance gains by adding synthetic backgrounds (Syn + real) show that blending both real and synthetic leads to the best performance, \eg +22.22 net improvement over Pure Real + cut paste in 10 shot regime.
Combining just 10-shot synthesized data with the 1,464 real training images achieves 68.38 mAP@50, a substantial +22.88 net improvement over using the 1,464 real training images alone.

On COCO, we observe a similar trend as VOC, \ie synthetic datasets provide a strong learning signal, while mixing both synthetic and real gives the most performance boost.

\subsection{More Baselines and Ablation Studies}
\label{sec:ablation}
\paragraph{Agnostic to backbones.}

\begin{figure}[ht]
  \centering
  \includegraphics[width=\linewidth]{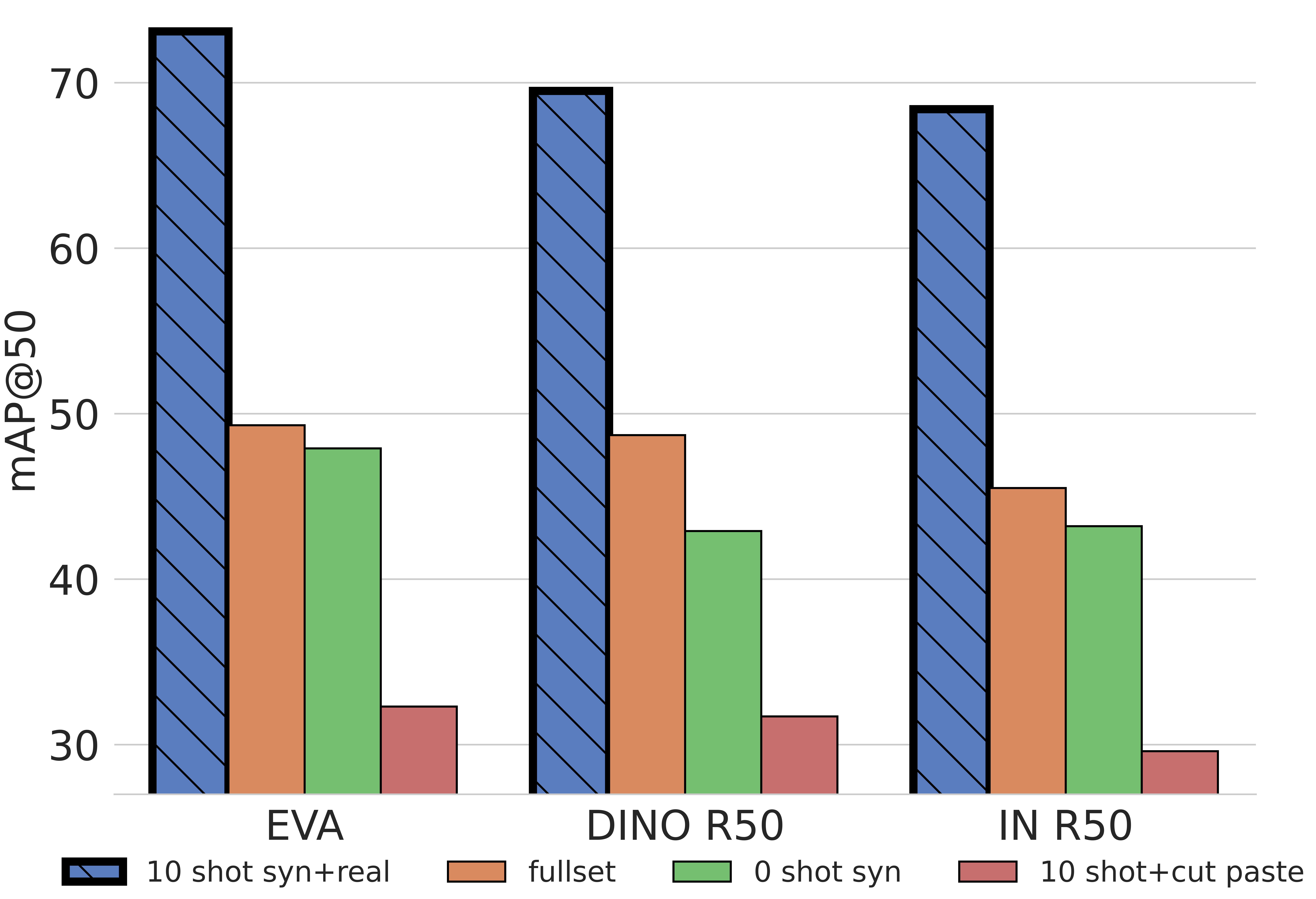}
  \caption{Our synthetic dataset generation is agnostic to different models and backbones.}
  \label{fig:ablation_on_backbone}
\end{figure}

We first show that our synthetic data generation pipeline is model-agnostic.
In \Cref{fig:ablation_on_backbone}, we present performance of two additional models: transformer-based EVA \cite{fang2023eva} and DINO self-supervised pretrained ResNet-50 \cite{caron2021emerging} on VOC.
We observe a similar trend across the model choice:
With only 10 shot exemplars, our approach can outperform fullest (pure real), which is 7x larger.
On the other hand, the model trained with our 0-shot-generated dataset can significantly surpass the best model available for training on 10 shot with cut-and-paste augmentation.

\begin{figure}[ht]
  \centering
  \includegraphics[width=\linewidth]{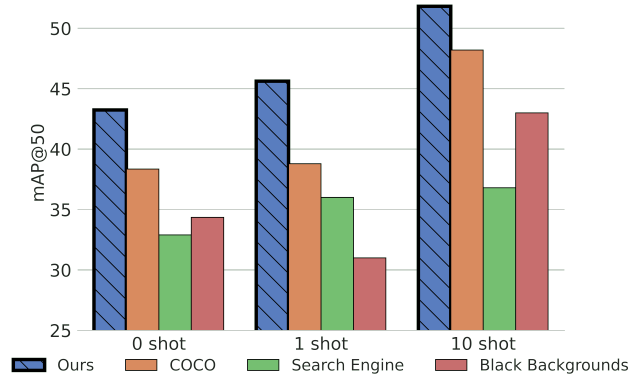}
  \caption{Contextual backgrounds generated by our approach provide valuable cues.}
  \label{fig:baselines on syn bg}
\end{figure}

\paragraph{Contextual backgrounds are crucial.}
We already demonstrated that in-context coherent backgrounds are beneficial in \Cref{sec:det}.
We further investigate what are the best contextual backgrounds.
As shown in \Cref{fig:baselines on syn bg}, on VOC we compare our synthetic contextual background from context-guided synthesis (\Cref{sub:bg_gen}) with three other contexts:
(1) Search Engine: substitute \syn with a search engine. Specifically, we directly collect backgrounds from Google search by using the same prompts as described in \Cref{sub:bg_gen};
(2) Other real datasets: use MS COCO dataset as background.
We randomly sample COCO images that contain only the remaining 60 classes so that they are disjoint from VOC 20 objects;
(3) Black background: replace each of the contextual backgrounds with pure black backgrounds.
Contextual backgrounds consistently outperform other baselines, suggesting that contextual cues in the backgrounds are important to learn a detection model.

\begin{figure}[ht]
\vspace{-10pt}
  \centering
  \includegraphics[width=\linewidth]{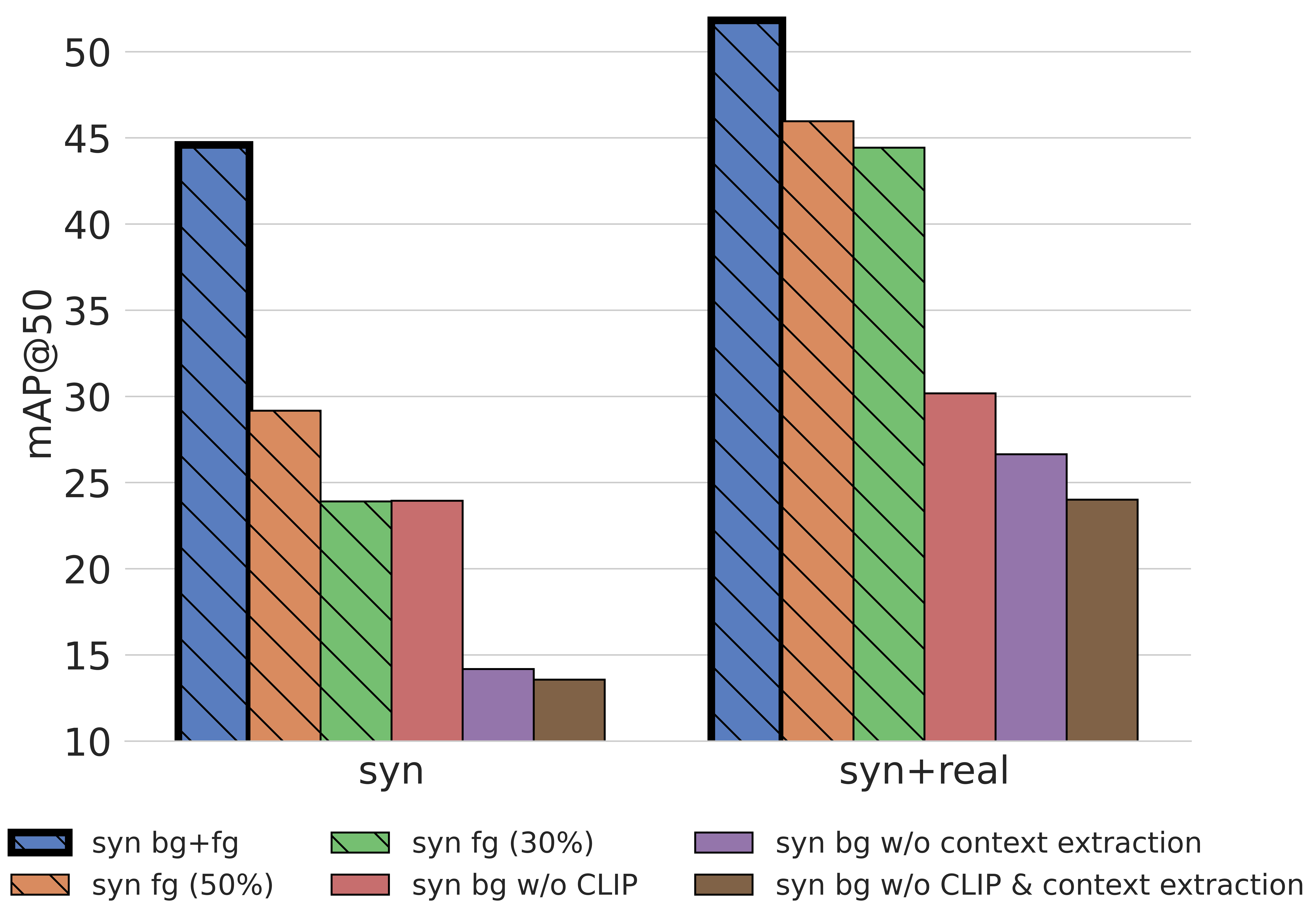}
  \caption{
 Our generated synthetic foregrounds (\texttt{fg}) are high-quality and diverse, and adding more helps.
 On the other hand, CLIP filtering and context extraction are crucial tricks to ensure the quality of synthetic backgrounds (\texttt{bg}).}
  \label{fig:ablation_on_syn_fg}
\end{figure}

\paragraph{CLIP and context extraction controls semantic quality and cleanness.}\label{sub:CLIP_is_beneficial}
We use CLIP \cite{radford2021learning} to filter and rank the synthesized context backgrounds (\Cref{sub:bg_gen}).
In \Cref{fig:ablation_on_syn_fg}, we train VOC object detection model without CLIP.
We observe at most -21.65 net decrease in performance, which implies that using CLIP as a variance reduction step is essential in pruning noisy or nonsensical backgrounds that might be generated by \syn.
We additionally ablate the effect of context extraction, as captions might contain interested objects. We found without extraction, interested objects contained in the captions will often be reflected in the synthetic images, and thus mislead the model during training as no annotation is provided in the pseudo-labeled dataset (\Cref{sub:cut paste}).

\noindent{\bf Extracting information from the generative model to enhance a discriminative model.}
Our foreground generation only requires class labels, so we can potentially generate as many foreground objects and their corresponding masks per class as we want.
In \Cref{fig:ablation_on_syn_fg}, we observe performance improvement as the number of foreground objects increases.

\paragraph{Robustness to foreground extraction methods.}
We next empirically demonstrate that foreground images consist of easy-to-separate background images. To this end, we use two off-the-shelf image segmentation methods, Entity Segmentation \cite{qi2021open} and PP-Matting \cite{chen2022pp} to segment out the foreground objects. We use the segmented foreground to generate training data and show their results on the Pascal VOC object detection task in \Cref{tab:ablation_of_entseg}. We observe that we achieve similar mAP scores in both these settings, which demonstrates that our approach is generally robust to the selection of the image segmentation method.

\begin{table}[htpb]
    \centering
    \small
    \begin{tabular}{c|l|cc}
    \toprule[1pt]
        \textbf{\#CDI} & \textbf{Method} & \textbf{EntSeg} \cite{qi2021open} & \textbf{PP-Matting} \cite{chen2022pp} \\ \hline
        0 shot & Pure Syn  & 43.24 & 46.11  \\
        \hline
        \multirow{3}{*}{\shortstack[c]{$20 \cdot 1$\\(1 shot)}} &  Syn Fg & 37.97 & 42.82 \\
         &  Pure Syn & 44.24 & 44.84 \\
        &  Syn + real  & 45.62 & 46.71  \\
        \hline
        \multirow{3}{*}{\shortstack[c]{$20 \cdot 10$\\(10 shot)}} & Syn Fg & 48.14 & 47.78  \\
         &  Pure Syn & 45.12 & 43.01 \\
        &  Syn + real  & 51.82 & 52.39 \\
        \bottomrule[1pt]
    \end{tabular}
    \caption{
    Our approach is robust to foreground extraction methods.}
    \label{tab:ablation_of_entseg}
\end{table}

\paragraph{Mixing real data with synthetic data.}
\begin{figure}[b]
  \centering
  \includegraphics[width=0.6\linewidth]{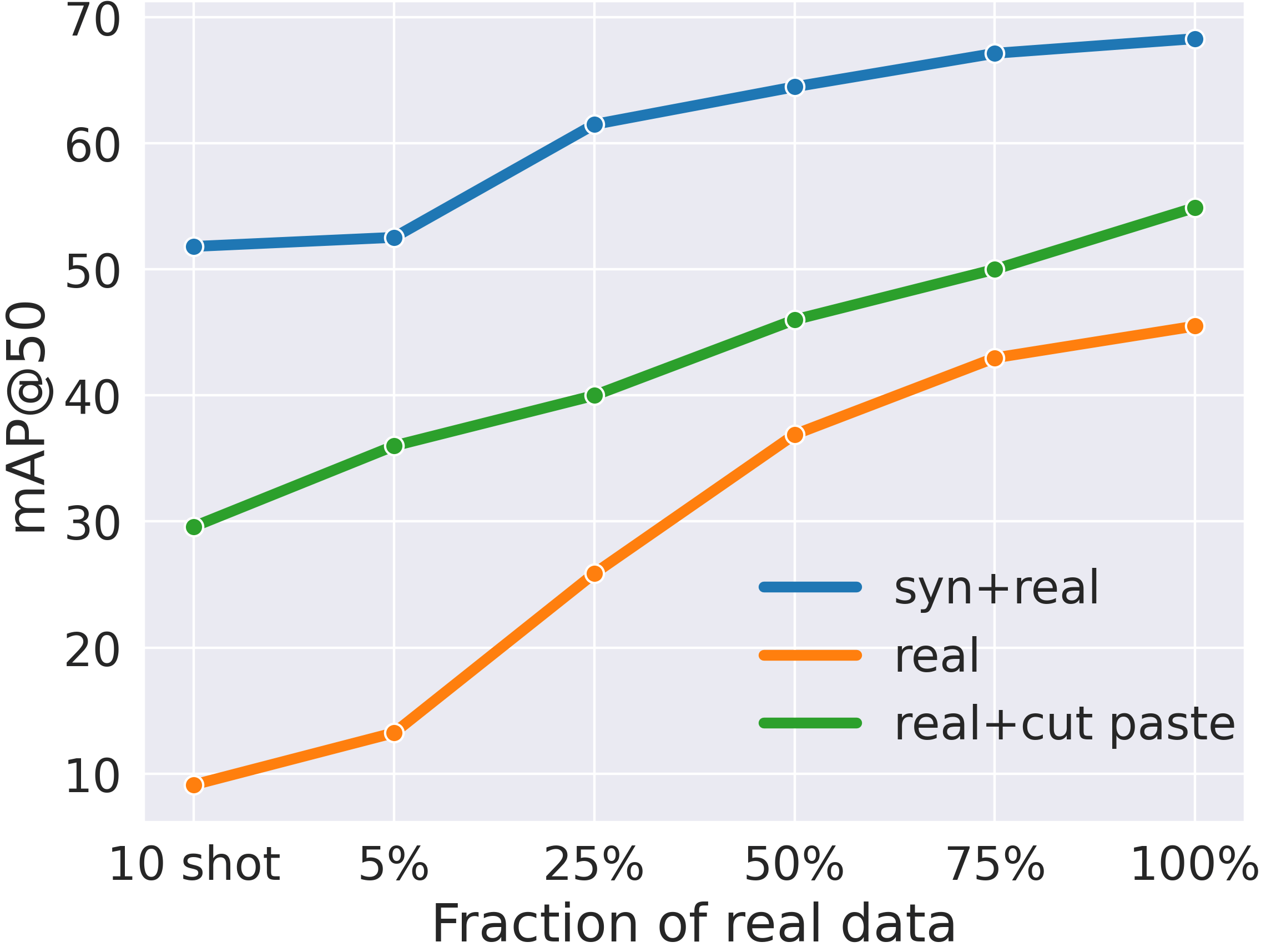}
  \caption{Mixing of real and synthetic data further improves downstream models.}
  \label{fig:real+syn}
\end{figure}
We show the effect of incorporating different percentages of real-world training images together with our synthesized images on Pascal VOC object detection.
In \Cref{fig:real+syn}, we experiment with adding additional $5, 25, 50, 75$ and $100$ percent of real images on top of the synthetic dataset generated in \Cref{sec:det}.
We observe strong performance gains compared to relying only on the same amount of real images or after applying cut-paste \cite{dwibedi2017cut}, \eg improving from 51.82 mAP to 68.38.
In \Cref{sec:generalization}, we show that such behavior is not limited to VOC dataset only.

\begin{figure}[t]
    \centering
    \includegraphics[width=\linewidth]{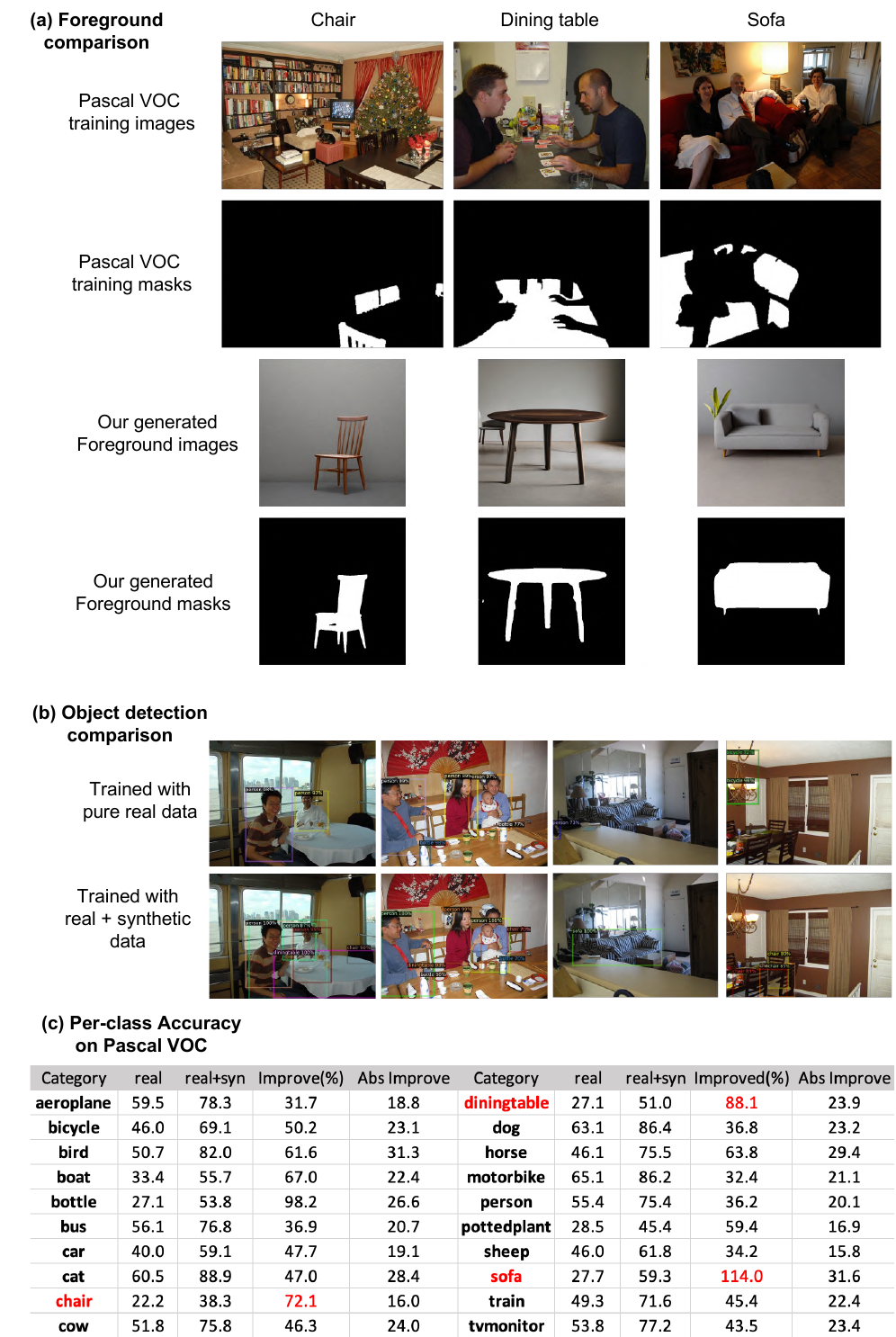}
    \caption{Synthetic data distribution complements the real data distribution. Our foreground generation helps even more on the highly occluded classes.}
    \label{fig:analysis}
\label{fig:analysis}
\end{figure}

\subsection{Synthetic data distribution complements the real data distribution.}
\label{sec:analysis}
Our results demonstrate that large \syn synthesis models can be used to generate high-quality training data for several large-scale problems. In order to analyze the behavior of our approach, we look at the object detection scores on VOC before and after adding synthetic data to full real training data.
Further, in \Cref{fig:analysis} (c), we calculate the per-class mAP and look at the relative improvement and overall scores.

We made two crucial observations. Firstly, there's a substantial relative increase in accuracy across all classes, with mAP50 values ranging from 30\% to 100\%. This suggests that the synthesized images significantly enhance the performance of downstream detection tasks. The marked improvement underscores how effectively the synthetic data distribution complements the real data distribution.
Secondly, to delve deeper into the synergy between synthetic and real data distributions, we examined the performance on a per-class basis. Notably, we witnessed marked enhancements in specific classes, such as the sofa, dining table, and chair.
We hypothesize that the significant improvement observed in indoor classes can be attributed to the generation of clean and unobstructed foreground objects using our approach. In real-world scenarios, these objects are typically occluded due to the presence of humans or other objects, as depicted in the first two rows of \Cref{fig:analysis} (a). Consequently, training images may exhibit similar characteristics, with a high degree of occlusion. However, our approach enables the generation of a diverse set of clean objects, which supplements the quality of the training data from the real world, as illustrated in the remaining two rows of \Cref{fig:analysis} (a).
This allows the training models to learn from both clean and occluded examples. 

Qualitative examples and the final object detection results on the test set are presented in \Cref{fig:analysis} (b), demonstrating that the model trained with a combination of synthetic and real data outperforms the model trained solely on real data, particularly for highly occluded categories.

\subsection{Generalization to more tasks}
\label{sec:generalization}
\paragraph{Instance Segmentation.}
\label{sec:instance_segmentation}
Our approach can generalize well to low-resource instance segmentation task on both Pascal VOC and COCO.
Following same settings as \Cref{sec:det}, in \Cref{tab:instance_seg}, we observe similar patterns across two datasets:
0 shot pure synthetic dataset yields strong performance while mixing real images further boosts the performance.

\begin{table*}[!htb]
    \begin{minipage}{.5\linewidth}
      \centering
      \small
    \begin{tabular}{c|l|cc}
    \toprule[1pt]
        \textbf{\#CDI} & \textbf{Method} & \textbf{mAP@50} & \textbf{mAP} \\ \hline
        \rotatebox[origin=c]{0}{0} &\graycell Pure Syn &\graycell 42.42 &\graycell 22.38 \\
        \hline
        \multirow{4}{*}{\rotatebox[origin=c]{90}{\shortstack[c]{$20 \cdot 1$\\(1 shot)}}} & Pure Real \cite{he2017mask}  & 0.00 &  0.00 \\
        & \ \ + cut paste \cite{dwibedi2017cut} & 1.27 & 0.88 \\
        &\graycell  Syn Fg &\graycell 42.23 &\graycell 21.80 \\
        &\graycell  Syn + real  &\graycell \textbf{46.74} \textcolor{red}{\small (+45.47)} &\graycell \textbf{24.81}  \textcolor{red}{\small (+23.93)}  \\
        \hline
        \multirow{4}{*}{\rotatebox[origin=c]{90}{\shortstack[c]{$20 \cdot 10$\\(10 shot)}}} & Pure Real \cite{he2017mask} & 5.21 &  2.09 \\
        & \ \ + cut paste \cite{dwibedi2017cut} & 24.50 & 9.24 \\
        &\graycell  Syn Fg  &\graycell 51.29 &\graycell 29.11 \\
        &\graycell  Syn + real  &\graycell \textbf{55.19} \textcolor{red}{\small (+30.69)} &\graycell \textbf{30.77}  \textcolor{red}{\small (+21.53)}  \\
        \bottomrule[1pt]
    \end{tabular}
    \end{minipage}%
    \begin{minipage}{.5\linewidth}
          \centering
      \small
    \begin{tabular}{c|l|cc}
    \toprule[1pt]
        \textbf{\#CDI} & \textbf{Method} & \textbf{mAP@50} & \textbf{mAP} \\ \hline
        \rotatebox[origin=c]{0}{0} &\graycell Pure Syn &\graycell 15.04 &\graycell 8.40  \\
        \hline
        \multirow{4}{*}{\rotatebox[origin=c]{90}{\shortstack[c]{$80 \cdot 1$\\(1 shot)}}} & Pure Real \cite{he2017mask}  & 0.03 & 0.00  \\
        & \ \ + cut paste \cite{dwibedi2017cut} & 3.83 & 1.87 \\
        &\graycell  Syn Fg &\graycell 15.10 &\graycell 8.23 \\
        &\graycell  Syn + real  &\graycell \textbf{17.56} \textcolor{red}{\small (+13.73)} &\graycell \textbf{9.12}  \textcolor{red}{\small (+7.25)}  \\
        \bottomrule[1pt]
    \end{tabular}
    \caption{Instance Segmentation for VOC (left) and COCO (above). Our methods generalize to other tasks and are competitive even in 1 shot.}
    \label{tab:instance_seg}
    \end{minipage}
\end{table*}

\paragraph{Object Instance Detection.}
\label{sec:kitchen_instance}
We evaluate our method on object instance detection tasks using three benchmarks: GMU-Kitchen \cite{georgakis2016multiview}, Active Vision \cite{ammirato2017dataset}, and YCB-video datasets \cite{calli2015ycb}.
For a fair comparison with \cite{dwibedi2017cut}, we instead use the object instance masks provided with the datasets and \emph{only synthesize backgrounds}.
In \Cref{table:active-vision-ycb-main} we compare our synthetic contextual backgrounds, generated from Ru-DALLE \cite{sberbank31:rudalle}, with context images from UW dataset \cite{georgakis2016multiview} following the setup of prior work \cite{dwibedi2017cut}.
Significant performance boosts indicate that our method is able to create congruent context images compared to real-world relevant images from public datasets.
Similar to \Cref{fig:analysis}, we report per-category mAP on GMU-Kitchen in \Cref{table:gmu-kitchen}.

Furthermore, similar to experiments done in \Cref{sec:ablation}, we investigate if the behavior of improved performance via blending real and synthetic can be transferred to this task.
In \Cref{table:vary_real} on the GMU kitchen, we 
use all synthetic backgrounds, but with various percentages of the real training images.
We observe that using only a subset of real-world data ($70 \%$) with our synthesized images achieves better performance than full (100\%) real-world data only. This suggests the advantages of our data generation approach saving the amount of human efforts required in labeling the real-world data significantly.
Further, we also observe that accuracy gradually improves from $78.3 \%$ to $91.4 \%$ as we increase the amount of real-world data.

\begin{figure*}[h]
\begin{center}
\includegraphics[width=\linewidth]{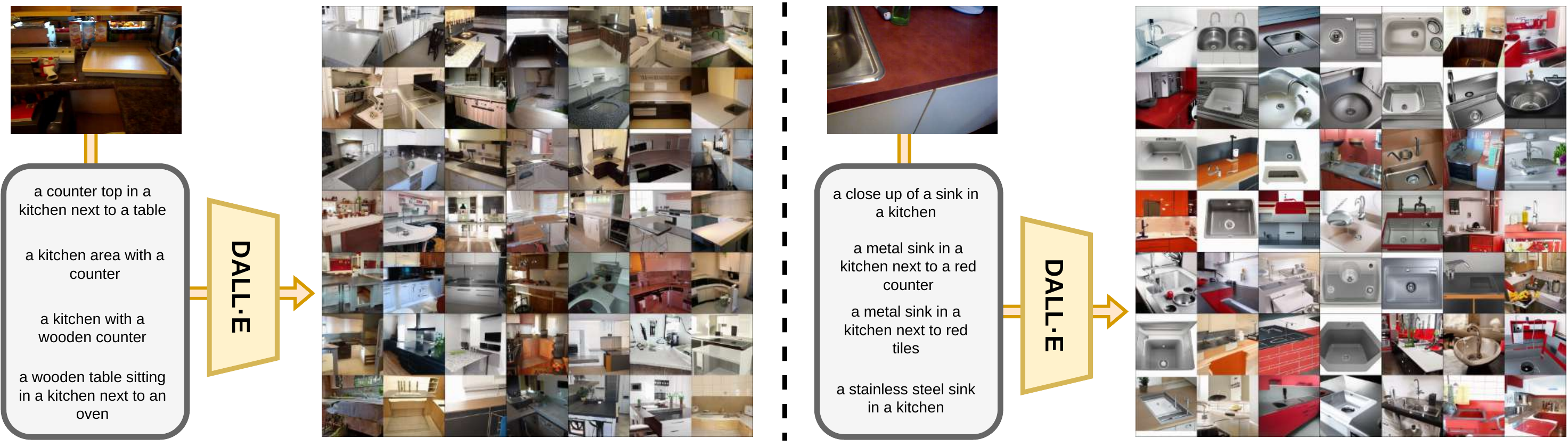}
\end{center}
   \caption{
   Contextual backgrounds generated from user-provided CDI.
   We note that even if the user provides \textit{as little as 1 CDI}, our approach can still generate large-scale coherent images.}
\label{fig:evidence for model able to generate a lot from single CDI }
\end{figure*}

\begin{table*}[!]
\begin{center}
\begin{tabular}{c|c|c|c|c|c|c|c|c|c|c|c|c}
\hline
    Dataset  & CC & CM & HB & HS & MR & NV1 & NV2 & PO & PS & Pbbq & RB & mAP  \\
\hline
  Syn (ours)  &  79.0  & 92.9 & 90.4 & 44.9 & 77.0 & 92.1 & 88.0 & 77.5 & 64.1 & 75.7  &  80.2 & 78.3  \\
  100\% Real  &  81.9  & 95.3 & 92.0 & 87.3 & 86.5 & \textbf{96.8} & 88.9 & \textbf{80.5} & 92.3 & 88.9  &  58.6 & 86.3  \\
  \hline
  Syn (ours) + 10\% Real  & 90.5 & 96.9 & 93.2 & 74.0 & 60.4 & 90.7 & 86.5 & 48.7 & 97.7 & 86.4 & 72.1 & 81.6 \\
  Syn (ours) + 40\% Real  & 91.8 & 97.4 & 94.5 & 84.9 & 75.1 & 90.7 & 78.6 & 52.1 & 96.9 & 87.6 & 77.9 & 84.3 \\
  Syn (ours) + 70\% Real  & 92.7 & 98.2 & 95.2 & \textbf{90.9} & 88.0 & 93.1 & 89.7 & 50.3 & 97.6 & 92.2 & 78.3 & 87.9 \\
  Syn (ours) + 100\% Real  & \textbf{94.4} & \textbf{98.2} & \textbf{95.2} & 90.7 & \textbf{92.5} & 94.1 & \textbf{93.0} & 72.8 & \textbf{98.3} & \textbf{98.7} & \textbf{79.8} & \textbf{91.4}  \\
\hline
\end{tabular}
\end{center}
\caption{We highlight that our synthesized data together with 70 $\%$ amount of real data achieves better performance than full (100 $\%$) set of real data only. This highlights the benefit of our approach in reducing total human efforts. Syn (ours) means ru-DALLE synthesized 1500 diverse images (use UW as CDI). Top row terms are: CC: Coca Cola, CM: Coffee mate, HB: honey bunches, HS: hunt's sauce, MR: mahatma rice, NV1: nature V1, NV2: nature V2, PO: palmolive orange, PS: pop secret, Pbbq: pringles bbg, RB: red bull.}
\label{table:vary_real}
\end{table*}

\begin{table}[b]
\begin{center}
\small
\begin{tabular}{c|c|c|c}
\toprule[1pt]
    \textbf{Dataset} & \textbf{GMU} & \textbf{Active Vision} & \textbf{YCB-video}    \\
\hline
  UW-Kitchen  & 76.1  & 22.6  & 38.3   \\
  DALL-E (ours) & \textbf{80.1}\textsubscript{\textcolor{red}{+4.0}}    & \textbf{25.8}\textsubscript{\textcolor{red}{+3.2}}  &  \textbf{45.5}\textsubscript{\textcolor{red}{+7.2}} \\
\bottomrule[1pt]
\end{tabular}
\end{center}
\caption{
Contextual synthetic backgrounds produced by our approach significantly enhance object instance detection accuracy across three datasets.
}
\label{table:active-vision-ycb-main}
\end{table}

\subsection{Compositionality in Synthetic Dataset}
\label{sec:compositionality}
\begin{table}[t]
\begin{center}
\footnotesize
\resizebox{\linewidth}{!}{
\begin{tabular}{c|c|c|c}
\toprule[1pt]
    \textbf{Dataset} & \textbf{Only CDI} & \textbf{No Intervention} & \textbf{After Intervention} \\
\hline
 Cartoon Kitchen & 11.2 &  70.0 & \textbf{76.7}\textsubscript{\textcolor{red}{+6.7}} \\
 Skeleton Kitchen  & 10.3 &  64.6 & \textbf{74.8}\textsubscript{\textcolor{red}{+10.2}} \\
 Objects in Kitchen & 9.4 &  71.8  & \textbf{77.0}\textsubscript{\textcolor{red}{+5.2}} \\
 Kitchens with Human  & 10.2 &  70.9 &  \textbf{76.9}\textsubscript{\textcolor{red}{+6.0}} \\
\bottomrule[1pt]
\end{tabular}
}
\end{center}
\caption{
Even if the user provides out-of-distribution CDI, our approach is able to produce a synthetic dataset tailored towards actual test distribution by in-domain intervention.
}
\label{table:compositional-mian}
\end{table}

As mentioned in \Cref{sub:bg_gen}, the compositional nature of our language-based context image generation allows us to {\em remove} noisy information, {\em add} relevant but missing information from the original textual description of the CDIs, or {\em change the style} of the received CDI to a more desired one.
For instance, the language description of a kitchen with people present in it may contain {``people''} as a distractor that may hamper the quality of the generated images and negatively affect the accuracy.
Using our pipeline, we can remove the distractor by detecting and removing it ({``people''}) from the caption before feeding them into \syn (\Cref{fig:compositionality}).

\noindent{\bf Contextual backgrounds from only one CDI.}
To demonstrate the compositionality of our approach, we extend the experimental setting from Object Instance Detection in \Cref{sec:ablation}: focusing on contextual background generation from \emph{only one} CDI on which GMU-Kitchen ground-truth objects are pasted.
In this case, CDI means the input image, but \emph{no longer in-domain}.
We consider 4 scenarios (Cartoon kitchen, Skeleton kitchen, objects in Kitchen and Kitchen with human), where provided CDIs are \emph{out of distribution} from the target domain, which is a real-world kitchen without humans.

There are two main challenges to (1) the conventional method struggles to learn effectively with as few as one training image (2) the provided training image is out-of-domain.
In \Cref{table:compositional-mian}, 
we demonstrate that our synthetic data generation pipeline can address both of these challenges. 
Specifically, when training solely on a synthetic dataset constructed by pasting objects onto a single CDI (\texttt{Only CDI}), performance across various settings is suboptimal. 
This underscores the inherent difficulties the model faces when learning from limited images and a narrow diversity.
However, 
our methodology can produce a vast collection of high-quality backgrounds that are contextually relevant.
The superior performance of the \texttt{No Intervention} results compared to the \texttt{Only CDI} results substantiates our hypothesis.
Qualitative results are presented in \Cref{fig:evidence for model able to generate a lot from single CDI } as additional empirical support that \syn is able to generate many relevant images from a single CDI using our approach.

\begin{figure*}[h]
\begin{center}
\includegraphics[width=0.8\linewidth]{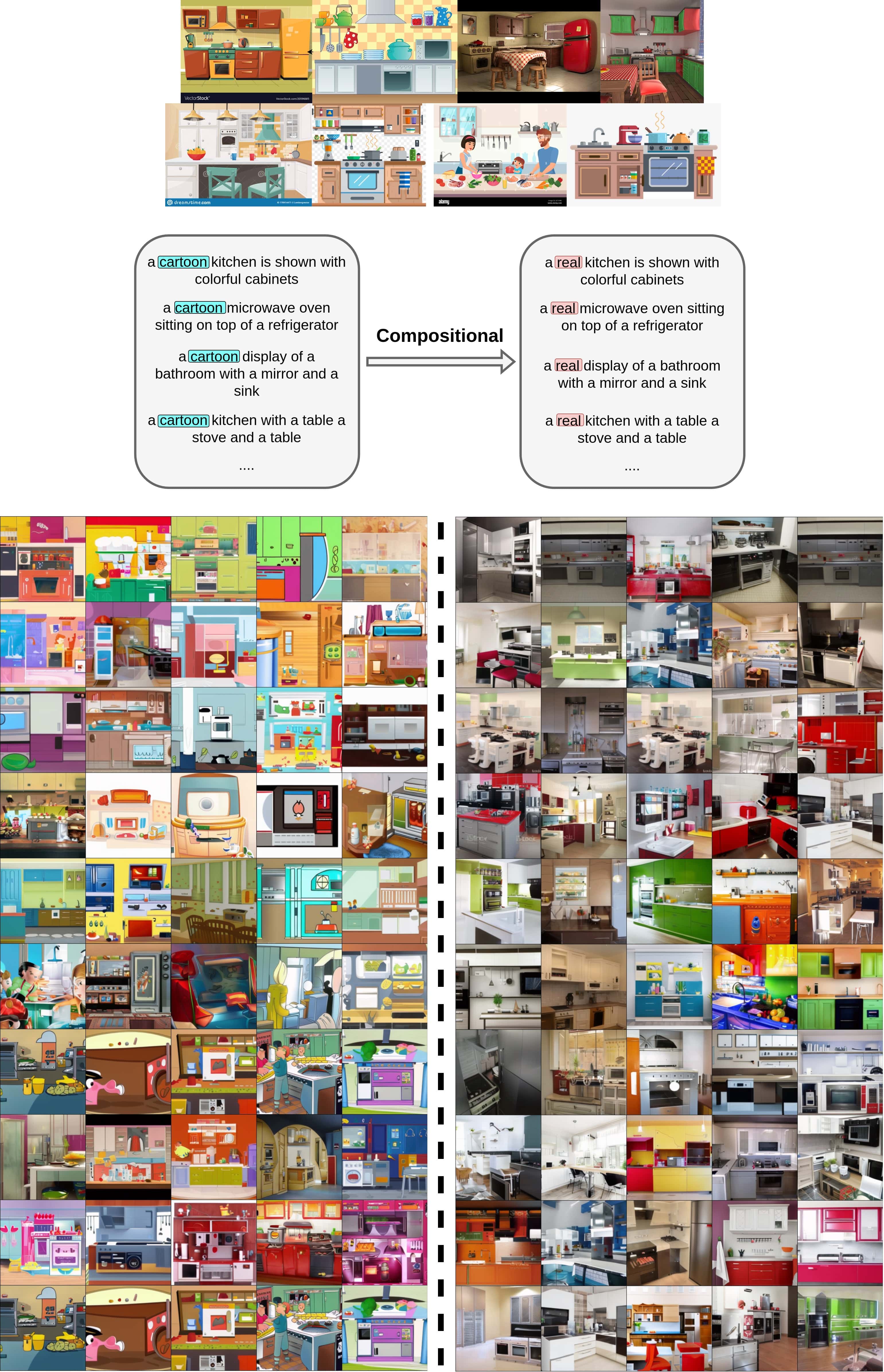}
\end{center}
   \caption{Visualization of context images generated by DALL-E for noisy and out-of-distribution CDIs. In the left column, inputs CDIs are cartoon kitchen which provide noise information about environment. The compositional property allow us \textbf{change the style} from cartoon to real and generate high-quality context images (right column).}
\label{fig:cartoon_appendix}
\end{figure*}

\begin{figure*}[h]
\begin{center}
\includegraphics[width=0.8\linewidth]{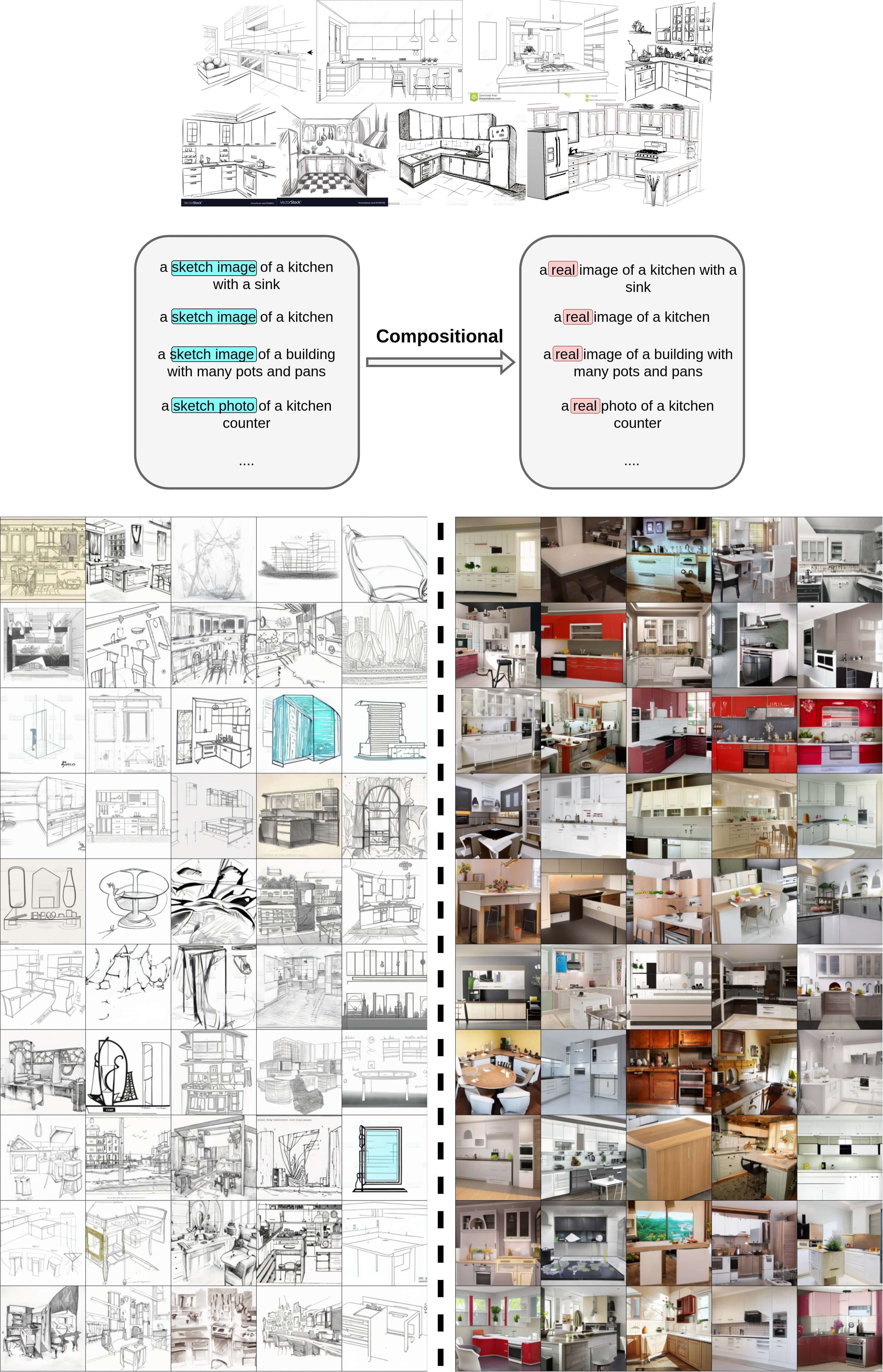}
\end{center}
   \caption{Visualization of context images generated by DALL-E for noisy and out-of-distribution CDIs. In the left column, inputs CDIs are sketch kitchen which provide noise information about environment. The compositional property allow us \textbf{change the style} from sketch to real and generate high-quality context images (right column).}
\label{fig:sketch_appendix}
\end{figure*}

\begin{figure*}[h]
\begin{center}
\includegraphics[width=0.8\linewidth]{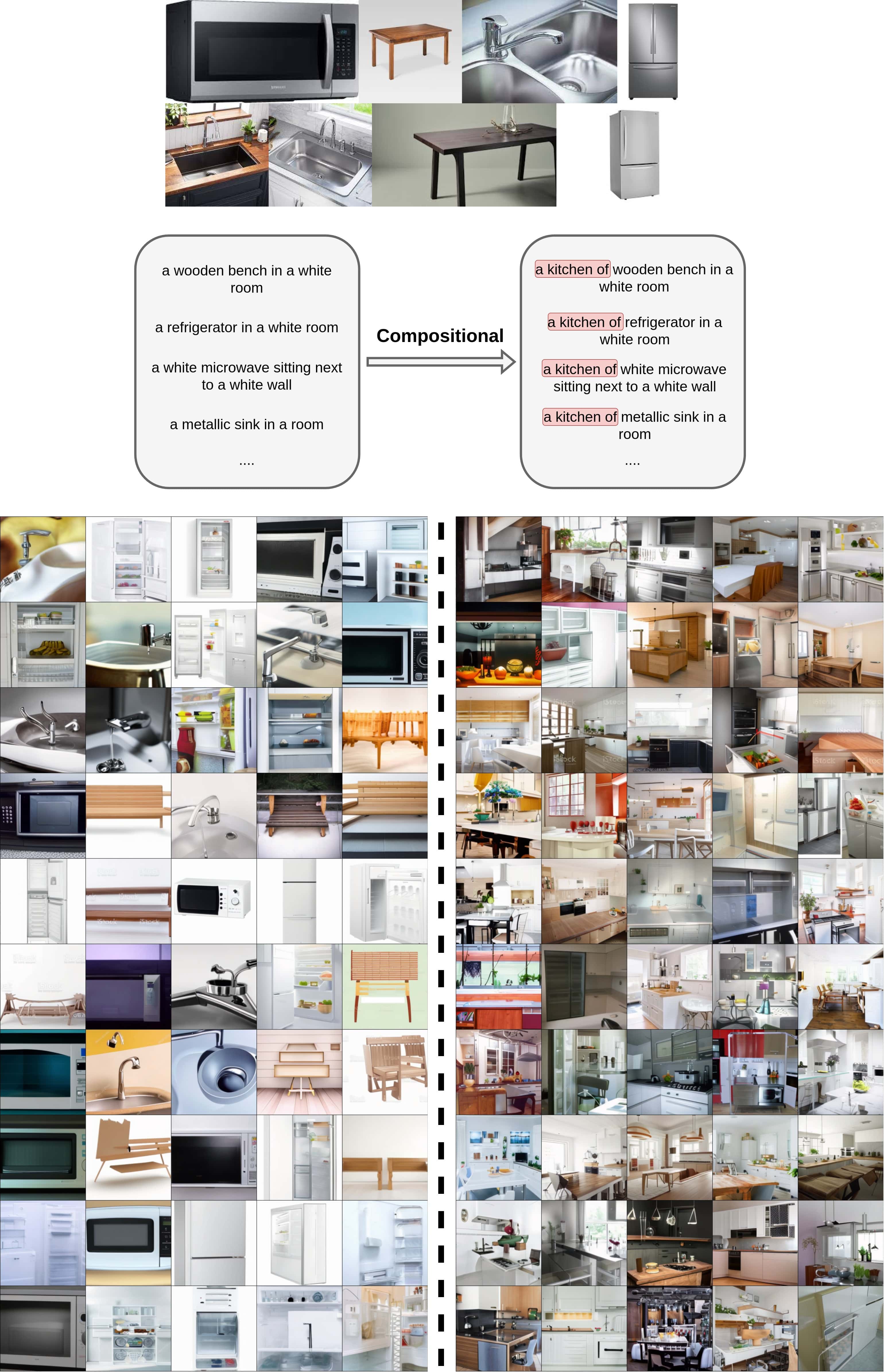}
\end{center}
   \caption{Visualization of context images generated by DALL-E for noisy and out-of-distribution CDIs. In the left column, inputs CDIs do not convey full knowledge about the context. The compositional property allow us \textbf{add} the context words to captions and generate high-quality context images (right column).}
\label{fig:object_appendix}
\end{figure*}

\begin{figure*}[h]
\begin{center}
\includegraphics[width=0.8\linewidth]{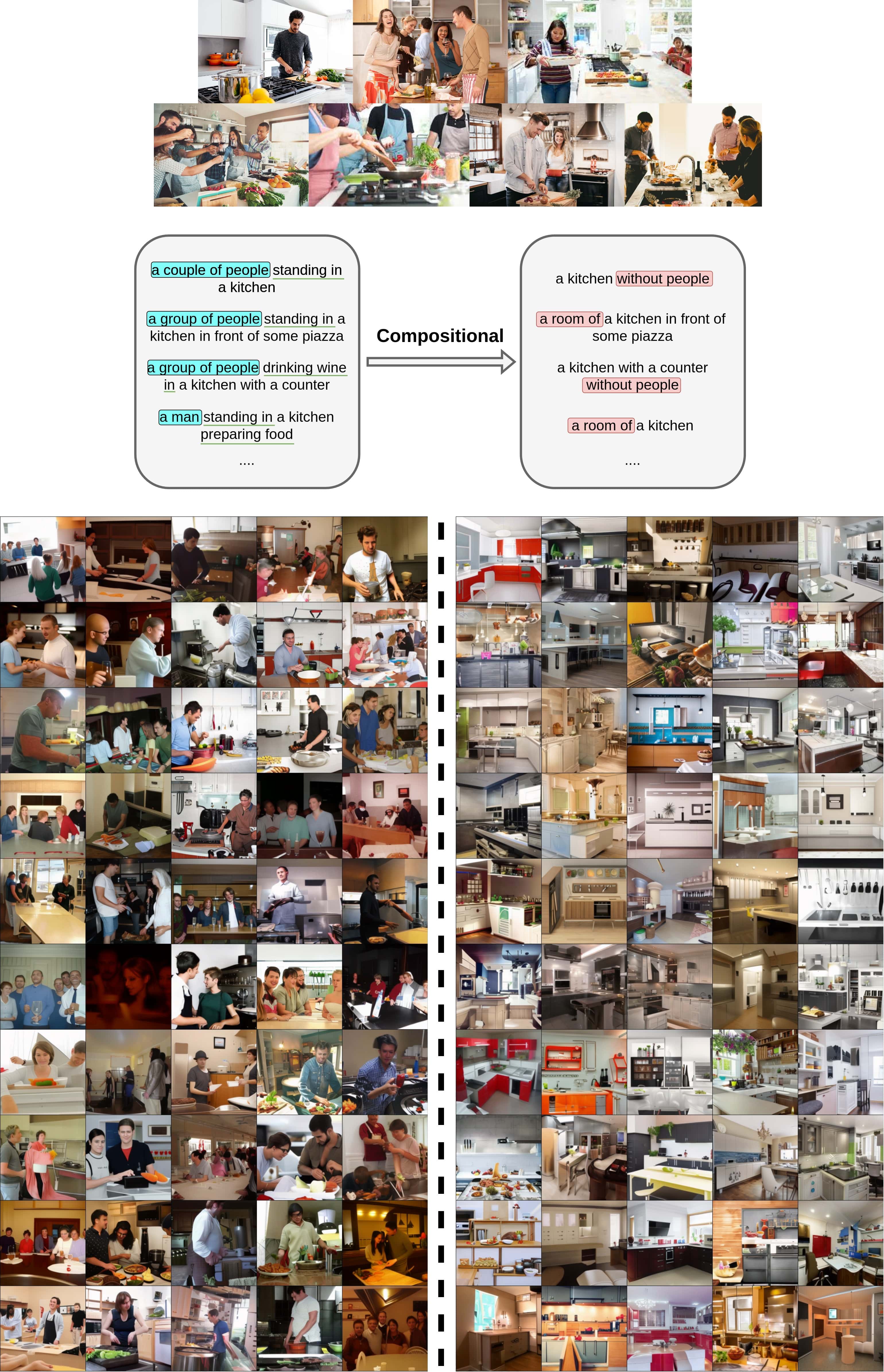}
\end{center}
   \caption{Visualization of context images generated by DALL-E for noisy and out-of-distribution CDIs. In the left column, inputs CDIs have distractor objects (people). The compositional property allow us \textbf{remove} the distractor words from captions and generate high-quality context images (right column). }
\label{fig:human_appendix}
\end{figure*}

\noindent{\bf Mitigate domain gap via language compositionality.}
While directly applying our approach leads to a significant performance improvement across all four scenarios, there remains room for enhancement due to the existing domain gap. Given that the contextual backgrounds are generated using augmented captions, \ie operating in the language space, in-domain intervention becomes feasible.
Specifically, we have the flexibility to add, remove, or alter contextual words, thereby influencing the style of the generated images. Consider the following interventions for each of the four scenarios we tested:
\begin{itemize}
    \item (\Cref{fig:cartoon_appendix}) we substitute the word ``cartoon'' with ``real'' to produce real kitchen images to better match test-time distribution.
    \item (\Cref{fig:sketch_appendix}) the word ``sketch image'' is replaced with ``real.''
    \item (\Cref{fig:object_appendix}) user-provided CDI only contains the kitchen object, which makes the generated images focus on the specific objects predominantly. To align more closely with the test-time distribution, we add the word ``a kitchen of.''
    \item (\Cref{fig:human_appendix}) as the test-time distribution involves real kitchen images without human presence, we remove words such as ``a couple of people,'' and add ``without people.''
\end{itemize}
In \Cref{table:compositional-mian} we observe up to 10.2 performance gain of \texttt{After Intervention} over \texttt{No Intervention}, which demonstrates the effectiveness of intervention in bridging the domain gap.
Such interventions ensure a closer alignment between the synthesis distribution and the desired test distribution.
\vspace{-1mm}
\section{Conclusion}
\label{sec:conclusion}
\vspace{-1mm}

We have proposed a new paradigm to generate large-scale labeled data for object detection and segmentation tasks using large vision and language-based text-to-image synthesis frameworks. We demonstrate effortless labeled data generation on popular benchmarks for object detection tasks. Computer vision models trained using these data improve the performance over the models trained with large real data. Thus reducing the need for expensive human labeling process. We also highlight the compositional nature of our data generation approach on out-of-distribution and zero-shot data generation scenarios. We believe our approach opens door to democratizing computer vision models. 
%
%
\\
\noindent
\textbf{Limitations.} Since we rely on T2I synthesis models for data generation, we are limited by two issues. First, our approach does not provide control for illumination, viewpoints, object pose and other such data generation properties. Second, our current approach can not generate labelled data for 3D geometry tasks like 3D object pose estimation tasks. We leave these problems as interesting future works.  

\noindent{ \bf Acknowledgements}
This work was supported in part by Amazon ML Fellowship, C-BRIC (one of six centers in JUMP, a Semiconductor Research Corporation (SRC) program sponsored by DARPA), DARPA (HR00112190134) and the Army Research Office (W911NF2020053). The authors affirm that the views expressed herein are solely their own, and do not represent the views of the United States government or any agency thereof.

\clearpage
\newpage

{\small
\bibliographystyle{ieee_fullname}
\bibliography{main.bbl}
}

\clearpage
\newpage
\section*{
Appendix
}
\begin{figure*}[h]
\begin{center}
\includegraphics[width=1.0\linewidth]{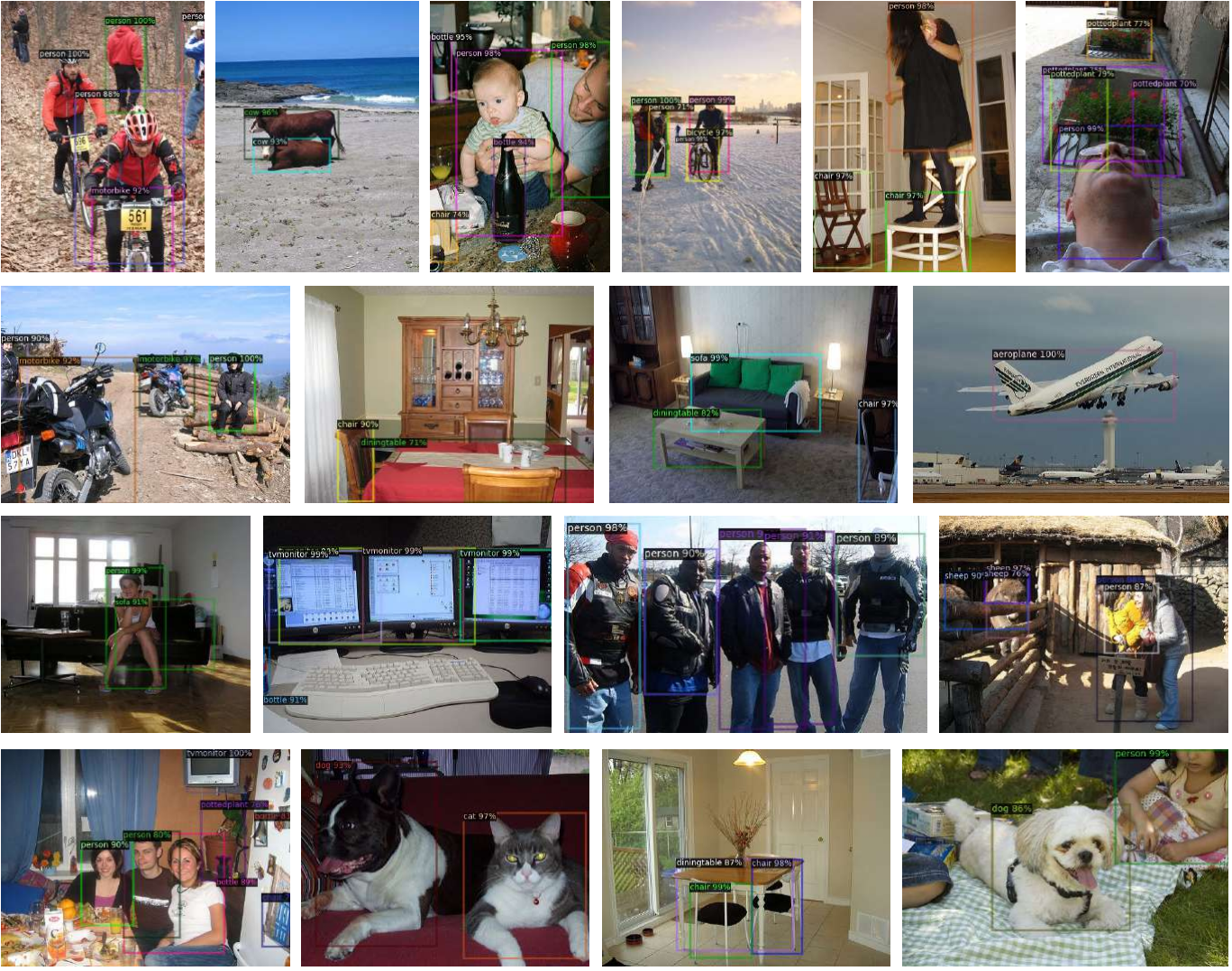} \\
\includegraphics[width=1.0\linewidth]{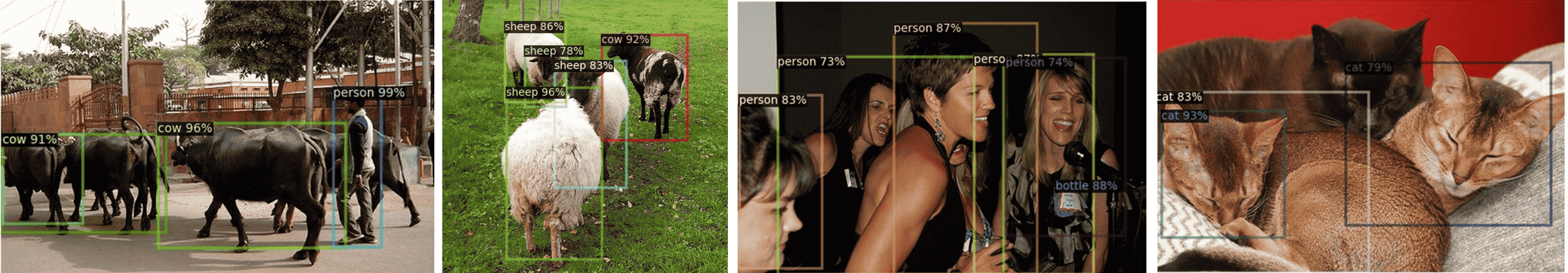}
\end{center}
   \caption{Prediction of Faster RCNN trained with 10 shot Pure Syn + Real in \Cref{tab:voc_object_detection}. The colorful box is the prediction of the model, together with the predicted class and confidence. We only show prediction with confidence $> 0.9$. The different color of the box indicates the different prediction of the box. Note how our model is able to handle complex real world images including occlusion, multiple objects, instances, etc.}
\label{fig:inference_vis_on_voc}
\end{figure*}

\begin{figure*}[h]
\begin{center}
\includegraphics[width=1.0\linewidth]{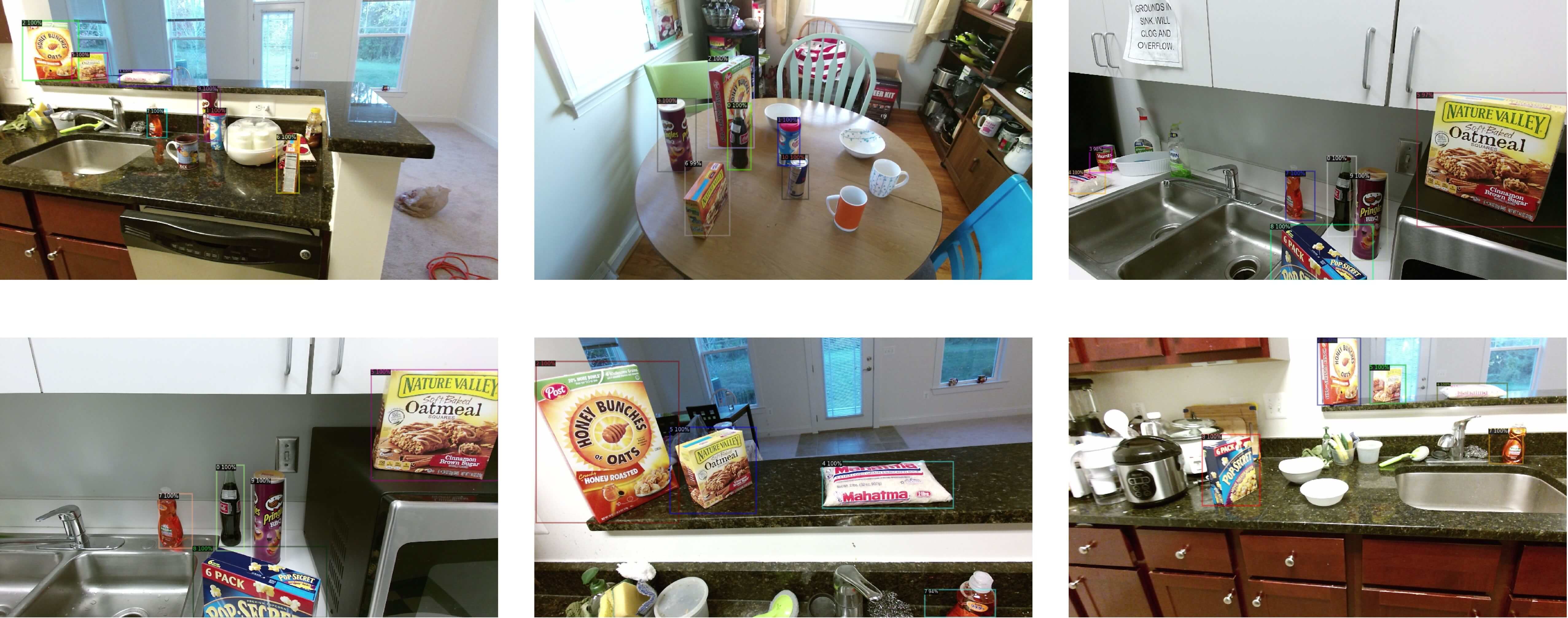}
\end{center}
   \caption{Qualitative detection results on GMU-Kitchen dataset dataset. The colorful box is the prediction of the model, together with the predicted class and confidence. We only show prediction with confidence $> 0.9$. The different color of the box indicates the different prediction of the box.}
\label{fig:inference_vis_on_gmu_kitchen}
\end{figure*}

We show inference results of Mask RCNN on VOC (\Cref{fig:inference_vis_on_voc}) and GMU-Kitchen (\Cref{fig:inference_vis_on_gmu_kitchen}). \Cref{fig:inference_vis_on_gmu_kitchen} shows the quantitative results on GMU-Kitchen dataset.

\begin{table*}
\begin{center}
\begin{tabular}{c|c|c|c|c|c|c|c|c|c|c|c|c|c}
\hline
    Dataset & \#CDI & CC & CM & HB & HS & MR & NV1 & NV2 & PO & PS & Pbbq & RB & mAP  \\
\hline
  Real GMU train& - & 81.9 & 95.3 & 92.0 & 87.3 & 86.5 & 96.8 & 88.9 & 80.5 & 92.3 & 88.9 & 58.6 & 86.3   \\
\hline
  Black & 1500 & 42.3  & 62.4 & 64.7 & 5.3 & 23.3 & 61.1 & 56.5 & 75.3 &  1.6 & 26.7 & 33.9  & 41.2  \\
  CDI &  10 & 51.4 & 26.4 & 2.1 & 12.2 & 12.1 & 0.4 & 0.1 & 1.0 & 0.1 & 29.8 & 30.0 & 15.0  \\
  Random (COCO) & 1500 & 50.7 & 80.1 & 77.5 & 15.3 & 32.2 & 81.7  & 87.9 & 71.7 & 66.8 & 59.0 & 68.5 & 62.8  \\
  Random (ru-DALLE) & 1500 & 64.8 & 86.9 & 78.7 & 49.2 & 62.2 & 84.8  & 83.8  & 72.6 & 70.9 & 57.2 & 24.1  & 66.8  \\
  UW-Kitchen &  1500  & 75.7  & 91.1 &  87.7 & 51.6 & 66.5 & 91.5 & 88.7  & 76.2 & 63.2 & 70.5 & 75.2 &  76.1  \\
  Syn (ours) & 1500 & 79.0 & 92.9 & 90.4 & 44.9 & 77.0 & 92.1 & 88.0 & \textbf{77.5} & 64.1 & 75.7 & 80.2 & 78.3  \\
  Syn (ours) & 2400 &  79.5  & 93.4 & 88.5 & 59.0 & 71.5 & 91.4 & 88.1 & 76.1 & 78.7 & 75.7  &  \textbf{80.6} & 80.1  \\
  \hline
  Syn (ours)+Real & 1500 & \textbf{94.4} & \textbf{98.2} & \textbf{95.2} & \textbf{90.7} & \textbf{92.5} & \textbf{94.1} & \textbf{93.0} & 72.8 & \textbf{98.3} & \textbf{98.7} & 79.8 & \textbf{91.4}  \\
\hline
\end{tabular}
\end{center}
\caption{Quantitative results on GMU-Kitchen dataset. We compare our approach with several prior approaches. Our approach achieves highest accuracy over the baselines. Combining GMU kitchen real training samples with our synthetic data yields the best results on this dataset.
}
\label{table:gmu-kitchen}
\end{table*}

\end{document}